\documentclass[]{academic_template}

\usepackage[toc,page,header]{appendix}

\usepackage{amssymb}
\usepackage{amsmath}

\usepackage{makecell}
\usepackage{algorithm}      % 算法浮动环境
\usepackage{algpseudocode}  % 算法伪代码语法

\usepackage{graphicx}

\usepackage{listings}
\usepackage{url}
\usepackage{booktabs}
\usepackage{wrapfig}

\usepackage{colortbl}
\usepackage{xcolor}

\usepackage{caption}
\usepackage{subcaption}
\usepackage{pifont}

\setlogoheight{14mm}   % logo size
\setlogospacing{5mm}
\setsjtublue % 设置logo为上交蓝，默认为上交红

\settitlerulethickness{3pt}  % title line size

\abstractboxon %显示边框
\setabstractframecolor{gray} %灰色边框
\setabstractbgcolor{gray!10} %浅灰色背景
\setlogotolineshift{5mm} % logo与标题行的距离

\settoprulethickness{2.5pt}  
\setbottomrulethickness{1.5pt} 

\makeatletter
\renewcommand\authorformat[2][]{\authorfont{\sffamily #2$^{#1}$}}
\makeatother

\title{Object Fidelity Diffusion for Remote Sensing Image Generation}

\author[1,2,*]{Ziqi Ye}
\author[3,*]{Shuran Ma}
\author[2]{Jie Yang}
\author[1]{Xiaoyi Yang}
\author[1]{Yi Yang}
\author[4]{Ziyang Gong}
\author[4, \dagger, \ddagger]{\linebreak[4] Xue Yang}
\author[1, \dagger]{Haipeng Wang}

\affiliation[1]{Fudan University}
\affiliation[2]{Shanghai Innovation Institute}
\affiliation[3]{Xidian University}
\affiliation[4]{Shanghai Jiao Tong University}

\contribution[*]{Equal Contribution}
\contribution[\dagger]{Corresponding Author}
\contribution[\ddagger]{Project Lead}

\abstract{
High-fidelity, controllable remote sensing layout-to-image generation is highly valuable for providing high-quality data for downstream object detection tasks. However, existing methods either rely on additional textual guidance, leading to geometric distortions, or require extra real-image references, limiting practical applicability. To address these challenges, we propose Object Fidelity Diffusion (OF-Diff), which leverages object layouts to extract structural shape priors and employs an online-distillation strategy to integrate complex image features. This allows the model to perform highly controllable, high-fidelity image generation at inference without relying on real-image references.
Furthermore, we introduce DDPO to fine-tune the diffusion process, making the generated remote sensing images more diverse and semantically consistent. Comprehensive experiments demonstrate that OF-Diff outperforms state-of-the-art methods in the remote sensing across key quality metrics. Notably, the performance of several polymorphic and small object classes shows significant improvement. For instance, the mAP increases by 8.3\%, 7.7\%, and 4.0\% for airplanes, ships, and vehicles, respectively.
}

\date{\today}
\checkdata[Code]{\url{https://github.com/VisionXLab/OF-Diff}}
\checkdata[Conference]{The Fourteenth International Conference on Learning Representations (ICLR 2026)}

\begin{document}
\maketitle

% 如果需要目录，取消下面的注释
% \newpage
% \tableofcontents
% \newpage

\section{Introduction}
\label{sec:intro}

% 这是一个包含多种元素的示例章节

% 图1
\begin{figure}[h]
\begin{center}
%\framebox[4.0in]{$\;$}
\includegraphics[width=\columnwidth]{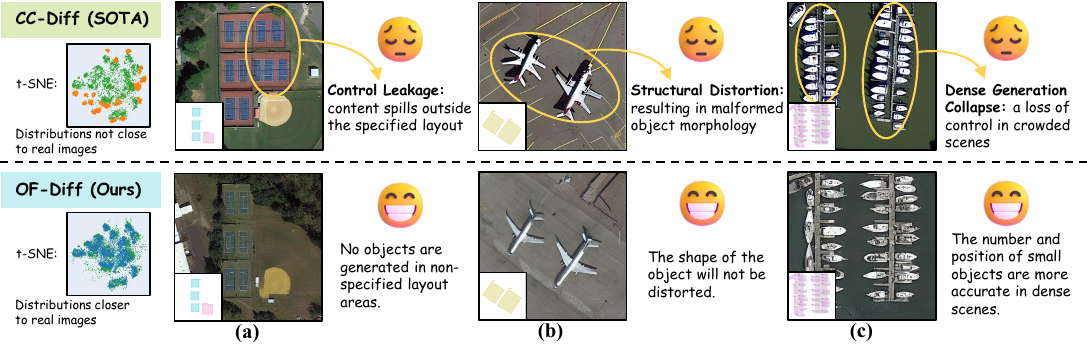}
\end{center}
\caption{Four critical failure modes in the State-of-the-Art (SOTA) method (CC-Diff): a distributional drift from real data, visualized by t-SNE; and (a) Control Leakage; (b) Structural Distortion; (c) Dense Generation Collapse. Our OF-Diff (2nd row) effectively resolves these issues.}
\label{fig:figure1}
\end{figure}

% 段落 1: 背景和动机
Synthesizing high-fidelity, spatially-controllable remote sensing (RS) images is a critical frontier for overcoming the data limitations that hinder downstream perception tasks like object detection \citep{yang2021r3det, zhang2020well, Yang_2019_ICCV}. Current RS generation methods, however, typically rely on either ambiguous text prompts \citep{khanna2023diffusionsat, sebaq2024rsdiff} or auxiliary conditions like semantic maps \citep{sebaq2024rsdiff, tang2024crs, gong2024crossearth, hu2025earth, jia2025can}. While visually plausible, such guidance is fundamentally disconnected from the instance-level ground truth, failing to provide the precise control necessary for effective data augmentation.

In contrast, Layout-to-Image (L2I) generation conditioned on object bounding boxes offers a more robust solution for precise spatial control. This paradigm has been extensively studied in the natural image domain—LayoutDiffusion \citep{zheng2023layoutdiffusion} treats it as a multi-modal fusion problem, GLIGEN \citep{li2023gligen} enables open-world generation through additional control signals, and ODGen \citep{zhu2024odgen} improves controllability by decoupling objects—yet its direct application to remote sensing (RS) imagery remains non-trivial due to expansive backgrounds, arbitrary object orientations, and densely packed scenes.

In RS layout-to-image generation, existing methods like AeroGen \citep{tang2025aerogen} and CC-Diff \citep{zhang2024cc} take different approaches. AeroGen, a coarse layout-conditioned model, suffers from limited spatial and shape control. In contrast, instance-level methods like CC-Diff achieve higher controllability and fidelity by referencing real instances, but this creates heavy dependence on the quality and quantity of real data, limiting generalization and flexibility.The images generated via CC-Diff diverge more markedly from the real remote sensing data distribution, aligning instead with the style characteristic of the model’s pre-training corpus. 
We summarize common failure modes (see Figure~\ref{fig:figure1}), including control leakage, structural distortion, dense generation collapse and feature-level mismatch.

\begin{figure}[h]
  \centering
  \includegraphics[width=1.0\textwidth]{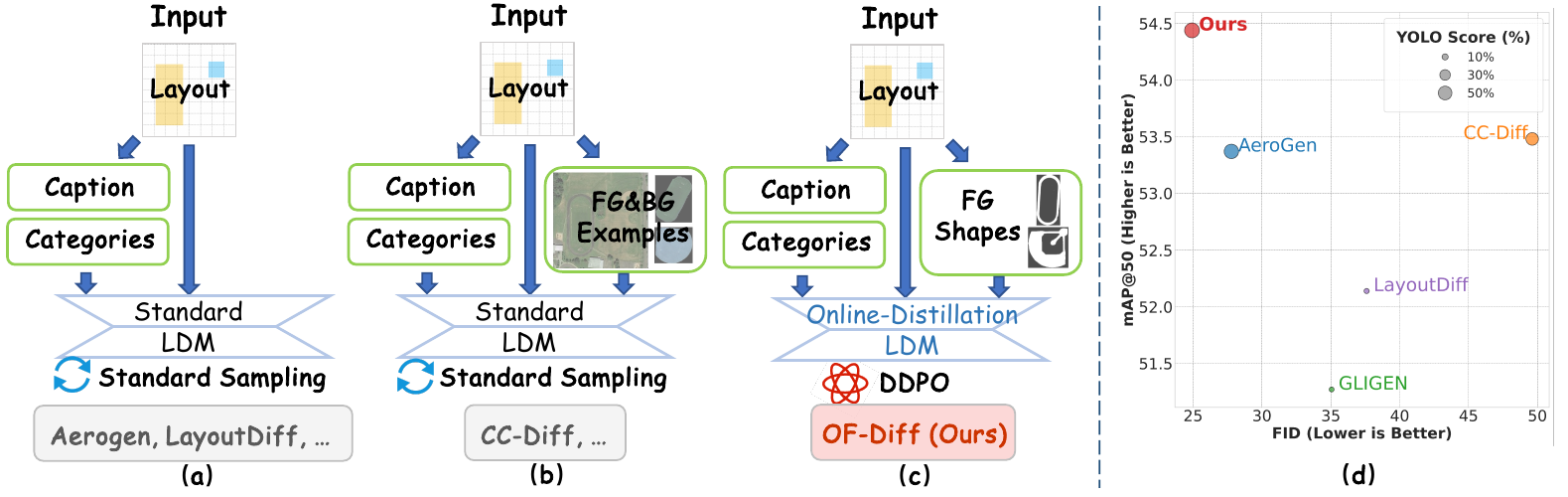}
  \caption{Comparison of OF-Diff with mainstream L2I methods. FG/BG stands for foreground/background. (a) Layout-conditioned baseline. (b) Added instance-based module, limited by quality/quantity of patches from ground truth. (c) OF-Diff enhances fidelity via shape extraction and DDPO, without patch reliance. (d) Results demonstrate superiority.}
  \label{fig:main_stream_and_bubble}
\end{figure}

These deficiencies significantly degrade the performance on object detection tasks, limiting their practical application in intelligent RS interpretation. In this paper, we introduce \textbf{O}bject \textbf{F}idelity \textbf{Diff}usion Model (OF-Diff). It is designed to improve the shape fidelity and layout consistency of object generation in RS images. As shown in Figure~\ref{fig:main_stream_and_bubble}, the existing L2I methods are mainly divided into two categories. The first is layout-conditioned baseline, as shown in Figure~\ref{fig:main_stream_and_bubble}(a), like Aerogen and LayoutDiffusion. The second is the method with instance-based module, as shown in Figure~\ref{fig:main_stream_and_bubble}(b), like CC-Diff. However, such methods require real instances and images as references during the sampling stage in order to generate high-quality synthetic images. In contrast, OF-Diff generates high-fidelity remote-sensing objects using only the foreground shape, and subsequently employs online-distillation to further align the outputs with real images, as shown in Figure~\ref{fig:main_stream_and_bubble}(c). In addition, it fine-tunes the diffusion with DDPO, effectively enhancing the performance of downstream tasks for the generated images. The results in Figure~\ref{fig:main_stream_and_bubble}(d) demonstrates the superiority of OF-Diff over other methods.
Our contributions are summarized as follows:

\begin{itemize}
        \item We introduce OF-Diff, an online-distillation controllable diffusion model with prior shape extraction, which improves generation fidelity while reducing reliance on real images, enhancing practical applicability.
        \item We propose a controllable generation pipeline that fine-tunes diffusion models with DDPO for remote sensing images, further boosting fidelity and diversity.
        \item Extensive experiments demonstrate that OF-Diff generates high-fidelity, layout- and shape-consistent images with dense objects, and serves as an effective enhancement for object detection tasks.
\end{itemize}

% 图2

\section{Related Work}
\label{gen_inst}

\subsection{Advances in Image Generation}
Diffusion models \citep{dhariwal2021diffusion, ho2020denoising, kingma2021variational} have increasingly replaced Generative Adversarial Networks (GANs) \citep{goodfellow2014generative, karras2021alias} and Variational Autoencoders (VAEs) \citep{kingma2013auto, rezende2014stochastic} in image synthesis tasks due to their training stability and superior output quality. Recent advances in efficient samplers, such as DDIM \citep{song2021denoising}, Euler \citep{karras2022elucidating}, and DPM-Solver \citep{lu2022dpm}, have further improved the practicality. Latent Diffusion Models (LDMs) \citep{Rombach_Blattmann_Lorenz_Esser_Ommer_2022}, which operate in low-dimensional latent spaces, significantly reduce computational costs while preserving visual fidelity. The success of models like DALL$\cdot$E2 \citep{ramesh2022hierarchical} and Imagen \citep{saharia2022photorealistic} demonstrates how this paradigm supports training on vast internet-scale datasets. As a result, diffusion-based approaches now provide a strong foundation for high-quality image generation.

\subsection{Layout-to-Image Generation}
Controllable image synthesis primarily includes text-to-image (T2I) and layout-to-image (L2I) generation. While T2I models \citep{nichol2022glide, ramesh2022hierarchical} achieve semantic alignment via textual prompts, L2I methods offer better spatial control. Recent works enhance layout conditioning through layout-as-modality designs \citep{zheng2023layoutdiffusion}, gated attention \citep{li2023gligen}, and instance-wise generation \citep{wang2024instancediffusion, Zhou_2024_CVPR}. However, these methods rely solely on coarse layout inputs (e.g., bounding boxes), which lack fine-grained shape information critical for synthesizing morphologically complex objects.

\subsection{Remote Sensing Image Synthesis}
Synthesizing high-fidelity training data is crucial for advancing remote sensing (RS) object detection, a field critical to numerous applications, but often hampered by the scarcity of extensively annotated datasets. Despite its necessity, most generative models for RS imagery, such as DiffusionSat \citep{khanna2023diffusionsat} and RSDiff \citep{sebaq2024rsdiff}, still rely on coarse semantic guidance. While other approaches leverage diverse control signals \citep{tang2024crs} like OpenStreetMaps \citep{espinosa2023generate}, they are generally not optimized for the bounding box format central to object detection. This naturally motivates L2I approaches including AeroGen \citep{tang2025aerogen} and CC-Diff \citep{zhang2024cc}, which have improved spatial accuracy and contextual consistency through layout-mask attention and FG/BG dual re-samplers. However, they suffer from limited controllability and heavy reliance on real data.

\section{Method}
\label{headings}

\subsection{Preliminary}
Diffusion models \citep{song2021denoising} aim to capture the underlying data distribution $ p(x) $ by iteratively reconstructing data from a noisy representation that is initially sampled from a standard normal distribution. Denoising Diffusion Probabilistic Models \citep{ho2020denoising} parameterize the model as the function \( \epsilon_{\theta}(x_t, t) \) to predict the noise component of the sample \( x_t \) at any time step \( t \). The training objective is to minimize the mean squared error (MSE) loss between the actual noise \( \epsilon \) and the predicted noise \( \epsilon_{\theta}(x_t, t) \):
\begin{equation}
\mathcal{L}=\mathbb{E}_{x_t, t, \epsilon \sim \mathcal{N}(\mathbf{0}, \mathbf{I})}\left[\left\|\epsilon_\theta\left(x_t, t\right)-\epsilon\right\|^2\right] .
\end{equation}

Stable Diffusion (SD) \citep{Rombach_Blattmann_Lorenz_Esser_Ommer_2022, qiu2025noise} utilizes a pre-trained VQ-VAE \cite{van2017neural} to encode images into a lower-dimensional latent space, performing training on the latent representation \( z_0 \). In the context of conditional generation, given a text prompt \( c_t \) and task-specific conditions \( c_f \), the diffusion training loss at time step \( t \) can be expressed as:
\begin{equation}
\mathcal{L}=\mathbb{E}_{z_t, t, c_t, c_f, \epsilon \sim \mathcal{N}(\mathbf{0}, \mathbf{I})}\left[\left\|\epsilon-\epsilon_\theta\left(z_t, t, c_t, c_f\right)\right\|^2\right] .
\end{equation} 

where $ \mathcal{L} $ represents the overall learning objective of the complete diffusion model. This objective function is explicitly applied during the fine-tuning of diffusion models in conjunction with ControlNet \citep{zhang2023adding}.

\begin{figure}[t]
  \centering
  \includegraphics[width=\textwidth]{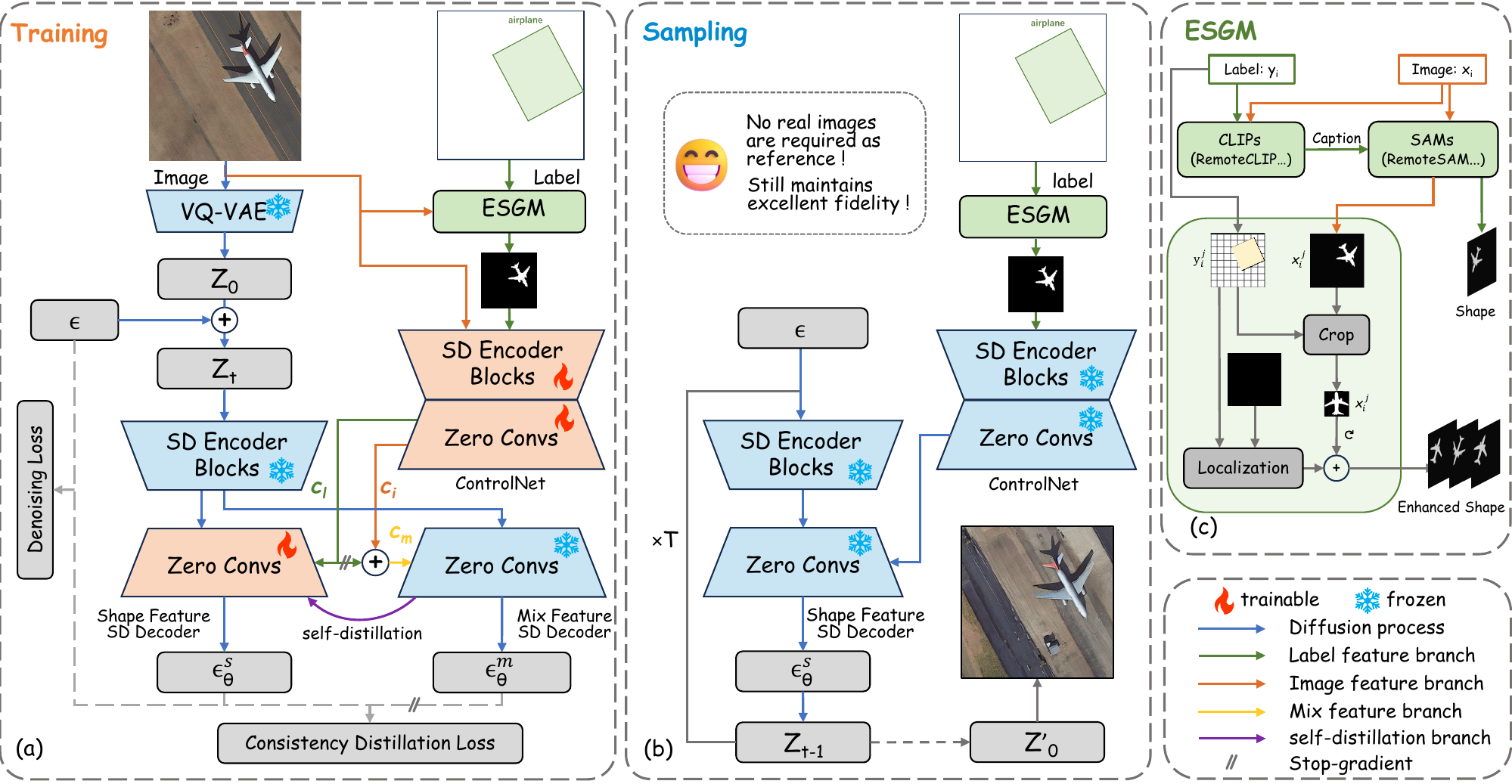}
  \caption{OF-Diff's overall architecture. (a) During training, object shape features extracted by ESGM and image features are processed by ControlNet, and the resulting information is used to update stable diffusion decoders via online-distillation. (b) During sampling, only the label and the shape feature stable diffusion decoder are used to generate synthetic images. (c) Architecture of the Enhanced Shape Generation Module (ESGM).}
  \label{fig:figure2}
\end{figure}

\subsection{Architecture of OF-Diff}

As illustrated in Figure~\ref{fig:figure2}(a), the training of OF-Diff requires both real images and their corresponding labels. \textbf{First}, for ControlNet, the real image and its label are processed by the Enhanced Shape Generation Module (ESGM) to extract the object mask. The image and mask are then fed into ControlNet to obtain the image feature \( c_i \) and the shape feature \( c_s \). To enrich the structural-only shape prior with richer appearance and contextual cues from the image, we combine them into a mix-feature \( c_m \), which will later serve as a teacher input in online-distillation. Concretely:

\begin{equation}
c_{\operatorname{m}}=\frac{n}{N} \cdot c_i+ \operatorname{sg}\left[c_s\right],
\end{equation}

where \( n \) denotes the current iteration number, and \( N \) is the total number of iterations. In order to enable the prediction conditioned on mix-feature to serve as a stable anchor point, to improve the morphological fidelity of the generation, we adopt a stop-gradient strategy \citep{Chen_He_2021} for \( c_s \) when calculating \( c_m \).

\textbf{Second}, for Stable Diffusion, the input image is first compressed into latent space features \( z_0 \) by a pre-trained VQ-VAE. Then, it is concatenated with Gaussian noise \( \epsilon \) to form \( z_t \). After passing through the SD encoder blocks, the feature \( Z_t \) is fed into a dual-decoder architecture. One branch, the shape-feature SD decoder, conditions on \( c_s \); the other, the mix-feature SD decoder, conditions on \( c_m \). Their reconstruction losses are defined as \(L_s\) and \(L_M\), respectively:

% The loss $ \mathcal{L}_s $ and $ \mathcal{L}_m $ of shape feature stable diffusion and mix feature stable diffusion are defined as:
\begin{equation}
\mathcal{L}_s=\mathbb{E}\left[\left\|\epsilon_\theta^s-\epsilon\right\|^2\right], \epsilon_\theta^s=\epsilon_\theta\left(z_t, t, c_t, c_s\right),
\end{equation}
\begin{equation}
\mathcal{L}_m=\mathbb{E}\left[\left\|\epsilon_\theta^m-\epsilon\right\|^2\right], \epsilon_\theta^m=\epsilon_\theta\left(z_t, t, c_t, c_m\right),
\end{equation}

\textbf{Third}, for online distillation, the mix-feature SD produces more accurate predictions thanks to its stronger image prior, but needs real images, limiting diversity. In contrast, the shape-feature SD supports arbitrary label control but risks converging to low-fidelity local minima. To reconcile these trade-offs, we propose an online-distillation framework with a consistency loss \(L_c\):

\begin{equation}
\mathcal{L}_c=\mathbb{E}\left[\left\|\epsilon_\theta^s-\operatorname{sg}\left[\epsilon_{\theta^{\prime}}^{\operatorname{m}}\right]\right\|^2\right].
\end{equation}

% The predicted noise \( \epsilon_{\theta^{\prime}}^{m} \) from the mix feature stable diffusion is more accurate than \( \epsilon_{\theta}^{s} \) predicted by the shape feature stable diffusion, owing to the additional image prior control. Nevertheless, during the sampling process, we do not utilize inputs with real images. Consequently, we employ self-distillation with stop-gradient operation so that the more accurate \( \epsilon_{\theta^{\prime}}^{m} \) can serve as an anchor, guiding the convergence trajectory of the masked diffusion towards high-fidelity local minima in the parameter space. 
% Ultimately, the loss employed to train OF-Diff is defined as:

Here, the prediction \( \epsilon_{\theta^{\prime}}^{m} \) from mix-feature SD decoder acts as a stop-gradient teacher signal, serving as an anchor to guide the prediction \( \epsilon_{\theta}^{s} \) from shape-feature SD decoder towards high-fidelity optima in parameter space. 

The overall training objective is therefore:

\begin{equation}
\mathcal{L}=\mathcal{L}_s+\mathcal{L}_m+\lambda\mathcal{L}_c,
\end{equation}

During the sampling phase, as illustrated in Figure~\ref{fig:figure2}(b), only the frozen ControlNet and the shape feature stable diffusion are utilized with arbitrary label prior control to synthesize RS images.

\subsection{Enhanced Shape Generation Module}
In natural imagery, perspective and scale changes prevent a unique geometric model for most objects. Conversely, remote-sensing objects display quasi-invariant shapes. For instance, courts are rectangular, chimneys and oil tanks circular, and airplanes bilaterally symmetric with a distinct nose and tail. This shape consistency enables the use of masks to impose strong controllability on image synthesis for remote sensing. To better leverage category labels for object shape extraction, we introduce the Enhanced Shape Generation Module (ESGM, see Figure~\ref{fig:figure2}(c)). During the training phase, ESGM uses paired images and labels to generate precise object masks. And at sampling time, it employs learned shape priors to synthesize diverse masks of object shape.

For the given image $x_i$ and its bounding box $y_i^{j}$ corresponding to category $j$ $(j \in [1, N])$, we first utilize the RemoteCLIP \citep{liu2024remoteclip} to generate a textual description of the object enclosed within the bounding box. With this description and the original image $x_i$, the RemoteSAM \citep{yao2025remotesam} then generates the corresponding shape masks \{$x_i^{j}$\}.

In the shape augmentation phase, each object mask $x_i^{j}$ is cropped by its bounding box $y_i^{j}$, randomly rotated, and placed back onto a blank canvas to produce a shape-enhanced mask. During training, ESGM uses real image shapes; at sampling, it selects enhanced shapes from a lightweight mask pool collected during or after training. In our experiments, we use masks generated during training.

\subsection{DDPO fine-tuning}
To enhance the diversity of the distribution of data generated by the fine-tuned model and maintain better consistency with the distribution of real images \cite{schulman2015trust, schulman2017proximal}, denoising diffusion policy optimization (DDPO) \cite{black2023training} is applied in the post-training of OF-Diff. DDPO regards the denoising process of the diffusion model as a multi-step Markov decision process (MDP) (for a detailed derivation, please refer to the Appendix~\ref{RL}). To optimize the policy $\pi(a_t \mid s_t)$ so as to maximize the cumulative reward $\mathbb{E}_{\tau \sim p(,\cdot \mid \pi)}!\Bigl[ \sum_{t=0}^{T} R(\mathbf{s}_t,\mathbf{a}_t) \Bigr]$, the gradient $\hat{g}$ is computed as follows:
\begin{equation}
\hat{g} = \mathbb{E}\!\biggl[ \sum_{t=0}^{T} \frac{p_{\theta}(\mathbf{x}_{t-1}\mid c,\ t,\ \mathbf{x}_{t})}{p_{\theta'}(\mathbf{x}_{t-1}\mid c,\ t,\ \mathbf{x}_{t})} \cdot r(\mathbf{x}_0, c) \cdot \nabla_{\theta} \log p_{\theta}(\mathbf{x}_{t-1}\mid c,\ t,\ \mathbf{x}_{t}) \biggr]
\end{equation}
\begin{equation}
r(\mathbf{x}_0, c) = \bigl( KNN(\mathbf{x}_0, \mathbf{x}_0) - \omega KL(\mathbf{x}_0, \mathbf{x}_0') \bigr)
\end{equation}

The reward functions based on K-Nearest Neighbor (KNN) and KL divergence are introduced, respectively, to optimize the diversity of generated data and the distribution consistency between generated data and real data. $\omega$ is the weight parameter, and $\mathbf{x}_0'$ is the real image in the dataset. Following standard practice, we compute the KNN in the low-dimensional embedding space of CLIP’s image encoder. The implementation details are shown in Appendix~\ref{RL}.

\section{Experiments}
\label{others}

\subsection{Experimental Settings}
\textbf{Datasets.}
\textbf{DIOR-R} \citep{cheng2022anchor}, the rotated variant of DIOR \citep{li2020object}, contains 20 categories annotated with oriented bounding boxes; we follow the official 1:1:2 split for training/validation/testing. \textbf{DOTA-v1.0} \citep{Xia_2018_CVPR} includes 15 categories featuring dense scenes and small objects. We crop the images from DOTA to $512\times512$ following MMRotate \citep{zhou2022mmrotate}, discarding those without valid objects. \textbf{HRSC2016}~\citep{hrsc2016} is a high‐resolution ship detection dataset with a multi‐level hierarchical taxonomy. We use the finest‐grained level, consisting of 26 detailed ship categories. The experiments of this dataset are reported in Appendix~\ref{hrsc2016_exp} Unless otherwise specified, we train the diffusion model on the trainset. For downstream detection, we use the trainset annotations as layout and mix generated samples with the real trainset, and report evaluation results on the testset.

\noindent\textbf{Implementation Details.} We train OF-Diff separately on each dataset (DIOR/DOTA), based on the Stable Diffusion 1.5 \citep{Rombach_2022_CVPR} pretrained model. Only the ControlNet and shape feature SD decoder are fine-tuned, while all other modules remain frozen. The weighting coefficient $\lambda$ of the consistency loss is set to 1, the k value in KNN is set to 50, and the weight $\omega$ of the KL divergence is set to 2. Training is performed using the AdamW optimizer with a learning rate of 1e-5. The global batch size is set to 64, and training runs for 100 epochs.

\noindent\textbf{Benchmark Methods.} 
We compare our method with state-of-the-art L2I generation models for both remote sensing (AeroGen \citep{tang2025aerogen}, CC-Diff \citep{zhang2024cc}), and natural images (LayoutDiffusion \citep{zheng2023layoutdiffusion}, GLIGEN  \citep{li2023gligen}). For a fair comparison, all models are re-trained using our dataset settings, following their official training details respectively.

\noindent\textbf{Evaluation Metrics. }
To more comprehensively evaluate the effectiveness of OF-Diff, we adopt a total of 13 metrics spanning 4 different evaluation aspects.
\begin{itemize}
	\item \textbf{Generation Fidelity. }
	We use \textbf{FID} \citep{heusel2017gans} and \textbf{KID} \citep{binkowski2018demystifying} to assess perceptual quality, along with \textbf{CMMD} \citep{jayasumana2024rethinking}, which measures CLIP feature distances between generated and real images to evaluate layout alignment.
	
	\item \textbf{Layout Consistency. } We report \textbf{CAS} \citep{ravuri2019classification} using a pretrained classifier to assess object recognizability, and \textbf{YOLOScore} by applying a pretrained Oriented R-CNN \citep{Xie_2021_ICCV} (w/. Swin Transformer backbone \citep{liu2021swin}, MMRotate) to generated images for instance-level consistency.
	
	\item \textbf{Shape Fidelity.} To assess the geometric quality of generated instances, we perform pairwise comparisons with ground-truth shapes. Each instance pair is cropped, resized to 64$\times$64, and converted to edge maps. We compute five metrics: \textbf{IoU}, \textbf{Dice}, Chamfer Distance (\textbf{CD}), Hausdorff Distance (\textbf{HD}), and \textbf{SSIM} \citep{wang2004image}. 
	
	\item \textbf{Downstream Utility. } We train a detector on mixed real and generated images and report \textbf{mAP$_{50}$}, \textbf{mAP$_{75}$}, and overall \textbf{mAP} on real test data using Oriented R-CNN (Swin backbone) with a batch size of 24 on 8$\times$NVIDIA 4090 GPUs.
\end{itemize}

\begin{figure}[t]
  \centering
  \includegraphics[width=\textwidth]{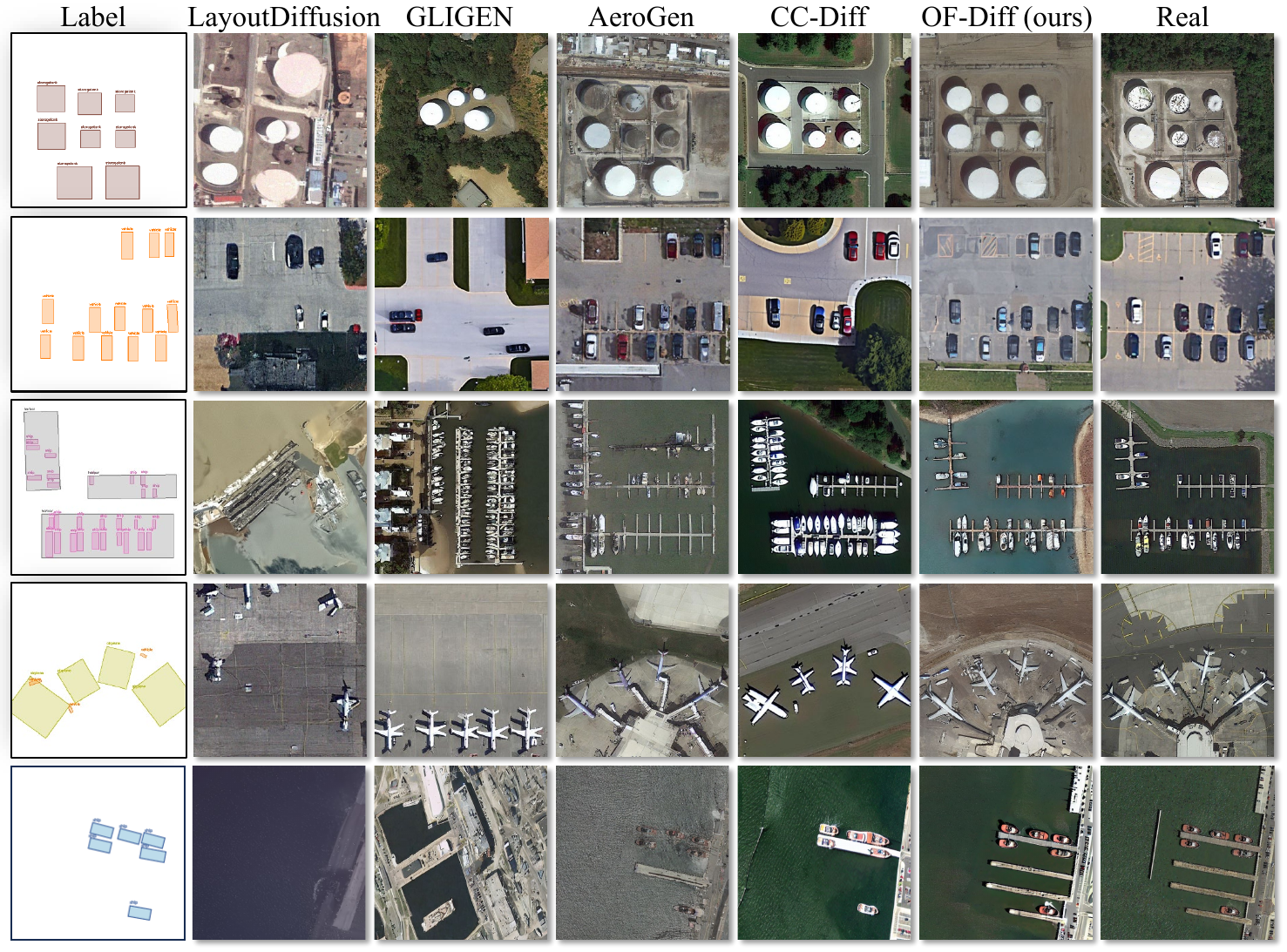}
  \caption{Qualitative results on DIOR, DOTA and HRSC2016. OF-Diff is more realistic and fidelity compared to other methods.}
  \label{fig:results_visualization}
\end{figure}

\subsection{Qualitative Results}
\textbf{Comparative Results.} 
Figure~\ref{fig:results_visualization} compares the generation results of OF-Diff with other methods. OF-Diff not only generates more realistic images but also has the best controllability. For instance, in the first two cases, OF-Diff successfully controlled the number and layout information of the generated objects. The third and fourth cases demonstrate the accuracy of OF-Diff in generating small targets, which other algorithms fail to do accurately. The last case shows the superiority of OF-Diff over other algorithms when generating objects with complex shapes such as airplane.
 
\noindent\textbf{Diversity Results.} 
The images generated by OF-Diff consistently present plausible textures and realistic object shapes, as shown in Figure~\ref{fig:diversity} in Appendix. For instance, airplanes rendered at different orientations maintain coherent semantic relationships with their surrounding environments. Even in small-object scenes (some of which are grayscale remote-sensing images from the DOTA dataset), OF-Diff can still generate visually faithful and geometrically accurate results.

\begin{table}[h]
	\centering
	% 9pt is allowed
	\caption{Quantitative comparison with SOTA methods on DIOR and DOTA. We evaluate performance on \textbf{generation fidelity} (FID, KID, CMMD), \textbf{layout consistency} (CAS, YOLOScore) and \textbf{trainability} (mAP). OF-Diff demonstrates superior overall performance.}
	\label{tab:main_results}
        \fontsize{7}{9}\selectfont 

	% 'l''c', left center
        \setlength{\tabcolsep}{3.2pt} % 压缩列间距（默认6pt）
	\renewcommand{\arraystretch}{1.1}
	\begin{tabular}{l | cccccc | cccccc}
		\Xhline{1pt} % head
		\multirow{2}{*}[-1pt]{\textbf{Method}} & \multicolumn{6}{c|}{\textbf{DIOR Dataset}} & \multicolumn{6}{c}{\textbf{DOTA Dataset}} \\ % [-6pt]手动调整上(+)下(-)
		\Xcline{2-13}{0.4pt} % 横线
		& FID$\downarrow$ & KID$\downarrow$ & CMMD$\downarrow$ & CAS$\uparrow$ & YOLOScore$\uparrow$ & mAP$_{50}$ & FID$\downarrow$ & KID$\downarrow$ & CMMD$\downarrow$ & CAS$\uparrow$ & YOLOScore$\uparrow$ & mAP$_{50}$ \\
		\Xcline{1-13}{0.4pt}
		LayoutDiff & 37.60 & 0.015 & \underline{0.447} & 70.32 & 7.01 & 52.14 & \underline{21.73} & \underline{0.015} & 0.288 & 77.56 & 21.43 & 66.75 \\
		GLIGEN & 35.06 & \textbf{0.010} & 0.622 & 76.41 & 6.51 & 51.27 & 39.79 & 0.026 & 0.357 & 76.19 & 15.58 & 66.10 \\
		AeroGen & \underline{27.78} & 0.013 & 0.563 & 81.69 & \underline{55.38} & 53.37 & 26.65 & 0.017 & 0.298 & \underline{81.91} & 44.85 & \underline{67.09} \\
		CC-Diff & 49.62 & 0.024 & 0.685 & \textbf{82.61} & 42.17 & \underline{53.48} & 32.40 & 0.019 & \underline{0.279} & 81.63 & \underline{49.62} & 66.52 \\
		\Xcline{1-13}{0.4pt}
		\textbf{Ours} & \textbf{24.92} & \underline{0.011} & \textbf{0.312} & \underline{82.55} & \textbf{58.99} & \textbf{54.44} & \textbf{20.84} & \textbf{0.014} & \textbf{0.271} & \textbf{83.79} & \textbf{55.68} & \textbf{67.89} \\
		\Xcline{1-13}{1pt}
	\end{tabular}
\end{table}

\begin{figure}[t]
  \centering
  \includegraphics[width=\columnwidth]{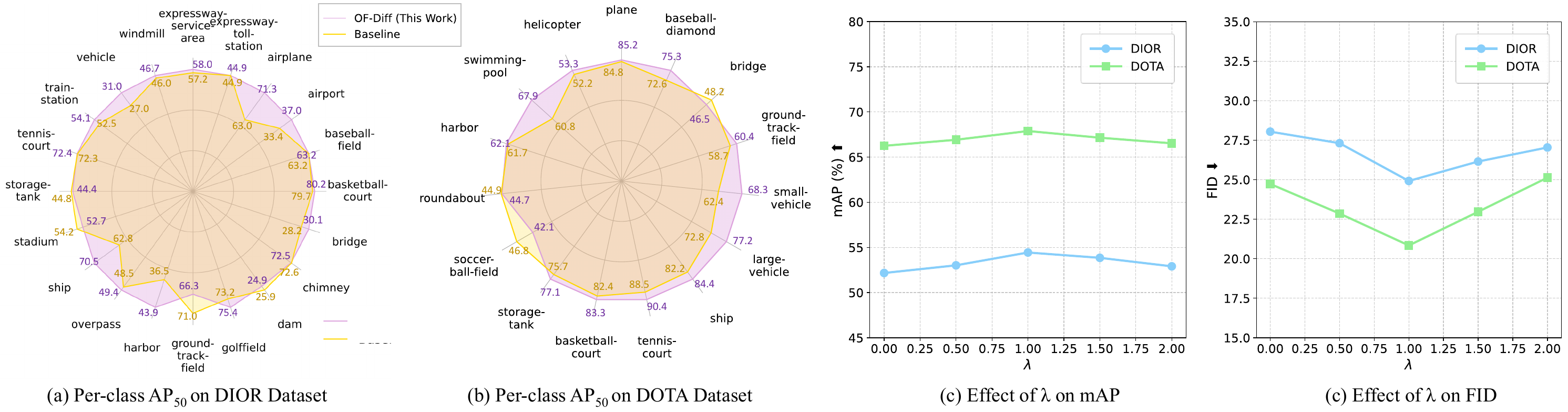}
  \caption{AP$_{50}$ on DIOR and DOTA.}
  \label{fig:radar_ap_dior_dota}
\end{figure}

% 形态学结果
\begin{table}[h]
	\centering
	% 9pt is allowed
	\caption{Object-Shape Fidelity on Canny Edge Maps. We measure the morphological similarity between generated and ground-truth instances by computing IoU, DICE, Chamfer Distance (CD), Hausdorff Distance (HD), and SSIM.}
	\label{tab:shape fidelity}
        \fontsize{7}{10}\selectfont 

	% 'l''c', left center
        \setlength{\tabcolsep}{7pt} % 压缩列间距（默认6pt）
	\renewcommand{\arraystretch}{1.1}
	\begin{tabular}{l | ccccc | ccccc}
		\Xhline{1pt} % head
		\multirow{2}{*}[-1pt]{\textbf{Method}} & \multicolumn{5}{c|}{\textbf{DIOR Dataset}} & \multicolumn{5}{c}{\textbf{DOTA Dataset}} \\ % [-6pt]手动调整上(+)下(-)
		\Xcline{2-11}{0.4pt} % 横线
		& IoU$\uparrow$ & Dice$\uparrow$ & CD$\downarrow$ & HD$\downarrow$ & SSIM$\uparrow$ & IoU$\uparrow$ & Dice$\uparrow$ & CD$\downarrow$ & HD$\downarrow$ & SSIM$\uparrow$ \\
		\Xcline{1-11}{0.4pt}
		LayoutDiff & 0.0497 & 0.0908 &	12.037 & 25.962 & 0.1667 & 0.0402 & 0.0748 & 15.229 & 30.202 & 0.2194 \\
		GLIGEN & 0.055 & 0.1002 & 12.257 & 25.850 & 0.1652 & 0.0645 & 0.1182 & 10.432 & 23.196 & 0.1967\\
		AeroGen & 0.0855 & 0.153 & 8.209 & 20.314 & 0.2142 & 0.0863 & 0.1536 & 8.1386 & 20.687 &	0.2261\\
		CC-Diff & 0.0891 & 0.1582 & 8.0909 & 20.066 & 0.1963 & 0.0692 & 0.1255 & 9.6226 & 21.247 & 0.2171\\
		\Xcline{1-11}{0.4pt}
		\textbf{Ours} & \textbf{0.1009} & \textbf{0.1763} & \textbf{7.6579} & \textbf{19.459} & \textbf{0.2691} &  \textbf{0.1205} & \textbf{0.2045} & \textbf{6.6317} & \textbf{17.311} & \textbf{0.2938} \\
		\Xcline{1-11}{1pt}
	\end{tabular}
	
\end{table}

%% 验证集结果
\begin{table}[t]
\centering
\caption{Quantitative comparison on the unknown layout dataset during training (DIOR Val).}
\label{tab:results_val}
    \fontsize{7}{10}\selectfont 
    % 'l''c', left center
    \setlength{\tabcolsep}{8pt} % 压缩列间距（默认6pt）
    \renewcommand{\arraystretch}{1.1}
\begin{tabular}{l | cccccccc}
\Xhline{1pt}
\multirow{2}{*}[0pt]{\textbf{Method}} & \multicolumn{8}{c}{\textbf{Unknown Layout during Training}} \\ 
\Xcline{2-9}{0.4pt}
& FID$\downarrow$ & KID$\downarrow$ & CMMD$\downarrow$ & CAS$\uparrow$ & \makecell{YOLO Score$\uparrow$} & mAP$\uparrow$ & mAP$_{50}$$\uparrow$ & mAP$_{75}$$\uparrow$\\
\Xcline{1-9}{0.4pt}
LayoutDiff & 44.58 & 0.018 & 0.539 & 29.34 & 10.37 & 30.41 & 53.07 & 32.07\\
GLIGEN      & 39.56 & \underline{0.013} & 0.444 & 66.36 & 2.13 & 30.06 & 52.68 & 31.29\\
AeroGen     & \underline{28.62} & \underline{0.013} & \underline{0.276} & \underline{80.78} & 46.36 & \underline{32.98} & \underline{55.11} & 34.26\\
CC-Diff     & 49.92 & 0.024 & 0.513 & 78.01 & \textbf{51.74} & 32.49 & 53.72 & \underline{35.39}\\
\Xcline{1-9}{0.4pt}
\textbf{Ours} & \textbf{24.18} & \textbf{0.012} & \textbf{0.271} & \textbf{83.34} & \underline{49.59} & \textbf{33.02} & \textbf{56.65} & \textbf{36.17}\\
\Xcline{1-9}{1pt}
\end{tabular}

\end{table}

%% 消融实验
\begin{table}[t]
\centering
\caption{Ablation study: impact of ESGM, Online-distillation \(L_c\), and DDPO on semantic consistency (CAS) and downstream trainability (YOLOScore and mAP$_{50}$). }
\label{tab:ablation_downstream tasks}
    \fontsize{7}{9}\selectfont 
    % 'l''c', left center
    \setlength{\tabcolsep}{8pt} % 压缩列间距（默认6pt）
    \renewcommand{\arraystretch}{1.1}
\begin{tabular}{ccc|cccccc}
\Xhline{1pt}
ESGM &  \(L_c\) & DDPO & FID $\downarrow$ & KID $\downarrow$ & CMMD $\downarrow$
& CAS $\uparrow$ & \makecell{YOLOScore $\uparrow$} & mAP$_{50}$ $\uparrow$ \\
\Xcline{1-9}{0.4pt}
\textcolor{gray}{\ding{55}} & \textcolor{gray}{\ding{55}} & \textcolor{gray}{\ding{55}}& 42.59 & 0.029 & 0.965 & 80.27 & 41.20 & 52.13 \\
\ding{51} & \textcolor{gray}{\ding{55}} & \textcolor{gray}{\ding{55}}& 24.87 & 0.012 & 0.428 & 82.16 & 55.08 & 52.76 \\
\textcolor{gray}{\ding{55}} & \ding{51} & \textcolor{gray}{\ding{55}}& 36.25 & 0.021 & 0.596 & 81.57 & 46.27 & 53.14 \\
\textcolor{gray}{\ding{55}} & \textcolor{gray}{\ding{55}} & \ding{51} & 41.26 & 0.027 & 0.815 & 81.06 & 42.53 & 53.41 \\
\ding{51} & \ding{51} & \textcolor{gray}{\ding{55}} & \underline{24.98} & \textbf{0.010} & \underline{0.313} & 82.30 & 57.83 & \underline{54.31} \\
\ding{51} & \textcolor{gray}{\ding{55}} & \ding{51} & 25.78 & 0.013 & 0.368 & \underline{82.37} & \underline{58.26} & 54.17 \\
\textcolor{gray}{\ding{55}} & \ding{51} & \ding{51} & 37.98 & 0.025 & 0.613 & 81.91 & 47.74 & 53.21 \\
\ding{51} & \ding{51} & \ding{51} & \textbf{24.92} & \underline{0.011} & \textbf{0.312} & \textbf{82.55} & \textbf{58.99} & \textbf{54.44} \\
\Xcline{1-9}{1pt}
\end{tabular}   

\end{table}

\subsection{Quantitative Results}
\noindent\textbf{Generation Fidelity and Consistency.} We compared OF-Diff with state-of-the-art generation methods in remote sensing, including layoutDiffusion \citep{zheng2023layoutdiffusion}, GLIGEN \citep{li2023gligen}, AeroGen \citep{tang2025aerogen}, and CC-Diff \citep{zhang2024cc}. The performance of these methods is reported in Table~\ref{tab:main_results}. OF-Diff achieved nearly the best performance in both generation fidelity metrics (FID, KID, CMMD) and layout consistency metrics, especially on the DOTA dataset. Additional results are available in the appendix~\ref{hrsc2016_exp} for the HRSC2016 dataset.

\noindent\textbf{Trainability of Object Detection.} 
Following the data enhancement protocol in \citep{Chen_Xie_Chen_Wang_Hong_Li_Yeung_2023}, we double the training samples using OF-Diff and assess detection results with the expanded dataset. As shown in Table~\ref{tab:mAP_results} in Appendix~\ref{sup_aesthe}, OF-Diff performs the best on both DIOR and DOTA with mAP improved by 2.2\% and 1.94\% compared to baseline,respectively. Notably, the performance of several polymorphic and small object classes shows significant improvement. According to Figure~\ref{fig:radar_ap_dior_dota} (a) and (b), the AP$_{50}$ increases by 8.3\%, 7.7\%, and 4.0\% for airplane, ship, and vehicle on DIOR, and 7.1\%, 5.9\% and 4.4\% for swimming pool, small vehicle, and large vehicle on DOTA.

\noindent\textbf{Object-Shape Fidelity.} 
We measure the morphological similarity between the generated instances and ground truth by calculating the Intersection over Union (IoU), DICE coefficient, Chamfer distance (CD), Hausdorff distance (HD), and Structural Similarity Index (SSIM), based on the Canny Edge Map. As shown in Table~\ref{tab:shape fidelity}, the results demonstrate that OF-Diff attains state-of-the-art performance in all evaluation metrics for object-shape fidelity. Specifically, we first convert the rotated bounding box (R-Box) to a horizontal bounding box (H-Box) and crop the instance with a 20\% padding to ensure the full object is captured. The cropped patches are then resized to $64\times64$ pixels, and their shapes are extracted using cv2.Canny. For a detailed qualitative comparison, Figure~\ref{fig:sup_edge_vis_combined} in Appendix~\ref{more_results} visualizes the instance patches and their corresponding edge maps from different methods. Each image set is ordered as follows: Ground Truth, OF-Diff, AeroGen, CC-Diff, GLIGEN, and LayoutDiff, demonstrating our method's superior ability to adhere to object shapes.

\noindent\textbf{Adaptability of Unknown Layout.} 
To evaluate robustness of these methods, we also generate images based on the unknown layouts during the training phase. According to Table~\ref{tab:results_val}, for unknown layout, OF-Diff performs well in terms of generation fidelity, layout consistency, and trainability. In downstream tasks, OF-Diff still delivers a 1.54\% mAP gain over the second-best method.

\noindent\textbf{The Detailed Results of Downstream.} 
Table~\ref{tab:DIOR-APs} and ~\ref{tab:DOTA-APs} in Appendix~\ref{more_results} report the average precision (AP) obtained by the competing generative methods over multiple categories in the downstream tasks. From Tables~\ref{tab:DIOR-APs}, it can be observed that OF-Diff (ours) achieves a clear advantage in several categories. For instance, OF-Diff achieves superior performance on Airplane (71.3\%), Golf Field (75.4\%) and Ship (70.5\%), with improvements of approximately 5\% to 10\% over the second-best method. For a few other categories, OF-Diff does not deliver the top AP, yet the gap to the best result remains marginal. Table~\ref{tab:DOTA-APs} shows that, on the DOTA dataset, OF-Diff obtains the highest AP in roughly half of all categories and still delivers notable gains in categories such as Small-vehicle (68.3\%), Ship (84.4\%) and Swimming-pool (67.9\%).

\subsection{Ablation Study}
We assessed the impact of different modules on image generation semantic consistency and downstream trainability by OF-Diff in Table~\ref{tab:ablation_downstream tasks}. We found that the images generated with captions are more in line with semantic consistency and human aesthetics, but the fidelity of these images decreases. This is equivalent to the data distribution deviating from the real dataset and being more inclined towards the data distribution during pre-training. We conduct human/GPT assessments and a fine-grained feature analysis in Appendix~\ref{sup_aesthe}, which collectively reveal the nature of this trade-off. Therefore, the ablation experiments for each module were conducted based on the absence of caption input. The contribution of each module to the enhancement of image generation fidelity is evaluated by incorporating additional components into the diffusion model with online-distillation. DDPO indicates whether to fine-tune the trained diffusion model through reinforcement learning. Results show that Enhanced Shape Generation Module (ESGM), Online-Distillation (\(L_c\)) and the DDPO based on KNN and KL Divergence effectively improve the performance metrics. Notably, ESGM can substantially improve the YOLOScore by over 10\%. In addition, we vary the weighting coefficient $\lambda$ in the consistency loss (Eq. 7) to assess its impact on mAP and FID. As shown in Figure~\ref{fig:radar_ap_dior_dota} (c) and (d), both metrics are optimal at $\lambda=1$.

\subsection{Discussion}
As shown in Figure~\ref{fig:w_o_caption} in Appendix~\ref{discuss}, the inclusion of additional captions as input has a significant impact on the outcomes of image generation. Specifically, incorporating captions enhances the aesthetic appeal of the generated images, resulting in richer and more visually pleasing color compositions. However, this improvement comes at a cost: similar to CC-Diff, it leads to a deviation of the generated data distribution from that of the original real data. In contrast, when no additional captions are provided as input, although the generated images may appear less aesthetically refined, their data distribution remains closer to that of real images. A user study from both human and GPT-5 in Table~\ref{tab:aethe-results} have confirmed this. Further analysis of the generated-image distribution and the impact of aesthetics on performance is provided in the Appendix~\ref{sup_aesthe}.

\subsection{Limitations}
The proposed OF-Diff injects object shape masks extracted from the image layout as controllable conditions into the diffusion model, which effectively enhances object fidelity and improves the generation of small objects. However, this also makes the model heavily dependent on the quality of the extracted shape masks. If the obtained masks are already distorted, the resulting generated images will likewise fail to achieve satisfactory quality. Therefore, in some scenarios, it is necessary to adopt more suitable mask extraction models (e.g., RemoteSAM, SAM2, SAM3) to obtain higher-quality masks. In future work, we will explore stronger object mask extraction and generation strategies to better ensure object-level generation quality while improving the texture details of the synthesized images. We will also investigate more advanced models and training paradigms to further promote the development of remote-sensing image generation.

\section{Conclusion}
Existing image generation methods struggle to precisely generate dense small objects and those with complex shapes, such as numerous small vehicles and airplanes in remote sensing images. To address this, we introduce OF-Diff, an online-distillation controllable diffusion model with prior shapes extraction. During the training phase, we extract the prior shape of objects to enhance controllability and use a online-distillation diffusion with parameter sharing to improve the model's learning ability for real images. Therefore, in the sampling phase, OF-Diff can generate images with high fidelity without real images as references. Finally, we fine-tune the diffusion by DDPO that combines KNN and KL divergence to make the synthesized images more realistic and consistent. Extensive experiments demonstrate the effectiveness and superiority of OF-Diff in generating small and difficult objects with complex structures and dense scenes in remote sensing.

\clearpage

\bibliographystyle{plainnat}
\bibliography{references}

% 如果需要附录，取消下面的注释
\clearpage
\beginappendix
% \appendix
\section{Appendix}
% \subsection{The Use of Large Language Models (LLMs)}
% Large Language Models (LLMs) were used in this work exclusively for language polishing and improving the clarity of writing. No LLM was used for generating scientific ideas, experimental design, or data analysis. All technical content, results, and interpretations presented in this paper are solely the work of the authors. The authors take full responsibility for the content of this paper, including any text refined with the assistance of LLMs.
\subsection{Reinforcement Learning Strategy}\label{RL}
The mapping relationship is defined as follows:

\begin{equation}
\pi(a_t \mid s_t) \triangleq p_{\theta}(\mathbf{x}_{t-1}\mid \mathbf{x}_t,c)
\end{equation}

\begin{equation}
P(s_{t+1}\mid s_t,a_t) \triangleq (\delta_{c},\,\delta_{t-1},\,\delta_{\mathbf{x}_{t-1}})
\end{equation}

\begin{equation}
\rho_0(s_0) \triangleq \bigl(p(c),\,\delta_{T},\,\mathcal{N}(0,I)\bigr)
\end{equation}

\begin{equation}
R(s_t,a_t) \triangleq
\begin{cases}
   r(\mathbf{x}_0,c), & \text{if } t = 0,\\
   0,        & \text{otherwise}.
\end{cases}
\end{equation}where $\delta_{z}$ denotes the Dirac delta distribution whose probability density is zero everywhere except at $z$. The symbols $s_t$ and $a_t$ represent the state and action at time $t$, respectively. Specifically, $s_t$ is defined as the tuple composed of the condition $c$, the time step $t$, and the noisy image $\mathbf{x}_t$ at that time, whereas $a_t$ is defined as the noisy image $\mathbf{x}_{t-1}$ from the preceding time step. The policy is denoted by $\pi(a_t \mid s_t)$, the transition kernel by $P(s_{t+1}\mid s_t,a_t)$, the initial state distribution by $\rho_0(s_0)$, and the reward function by $R(s_t,a_t)$.

%-----------------------------------------------------------------------------------------
% 添加原理解释
For detailed DDPO policy, we employ a ResNet101 pre-trained on ImageNet-1K as our feature extraction model, and utilize KNN and KL divergence to compute both the diversity among generated images and their similarity to real images. Let $X$ denote the set of generated images, $Y$ represent the real images, where $x_i \in X$, $y_j \in Y$, and $M$ is our feature extraction model.

The KNN reward is calculated as follows:
1) First, we extract features from $X$ using model $M$: $F_x = M(X)$.
2) For each feature vector $f_x^i \in F_x$, we compute its K-nearest neighbors among all feature vectors $f_x^j \in F_x$. The KNN reward for $x_i$ is the average of these K nearest neighbor distances, denoted as $KNN(f_x^i,F_x)$. In our implementation, we set K to 50.

The KL reward is calculated as follows:
1) We extract features from both $X$ and $Y$ using model $M$: $F_x = M(X)$ and $F_y = M(Y)$.
2) For each feature vector $f_x^i \in F_x$ and $f_y^j \in F_y$, we compute $KL(f_x^i,f_y^i)$ for each $i$, and use $-KL(f_x^i,f_y^i)$ as the KL reward for $x_i$.

In summary, the reward for a generated image $x_i$ is computed as:
\begin{equation}
r_x^i \triangleq KNN(f_x^i,F_x) - w KL(f_x^i,f_y^i)
\end{equation}

% 结束DDPO原理解释
%-----------------------------------------------------------------------------------------

\begin{figure}[t]
  \centering
  \includegraphics[width=0.8\columnwidth]{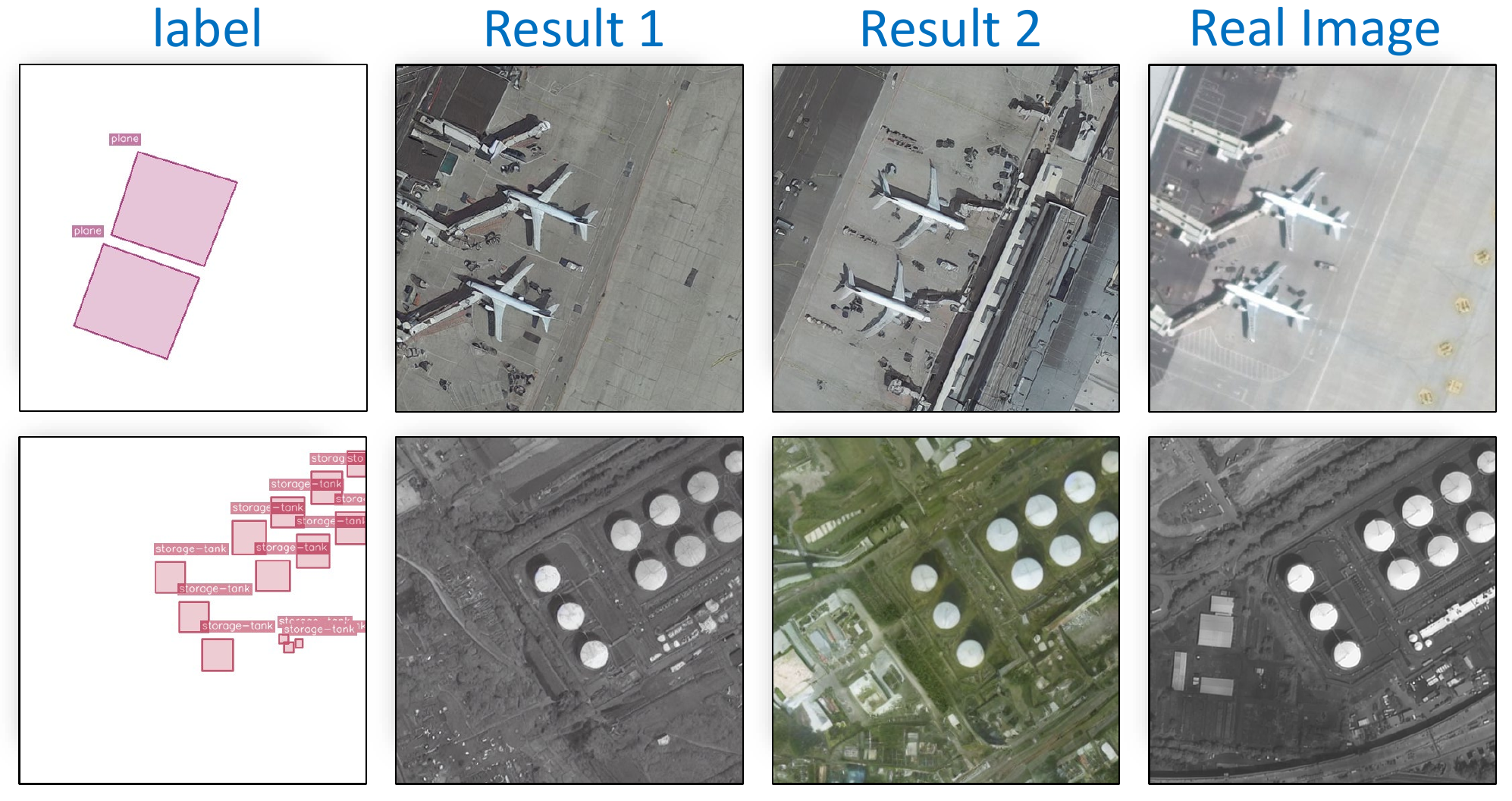}
  \caption{The diversity of different results from the same OF-Diff model.}
  \label{fig:diversity}
\end{figure} 

\begin{figure}[t]
  \centering
  \includegraphics[width=0.9\columnwidth]{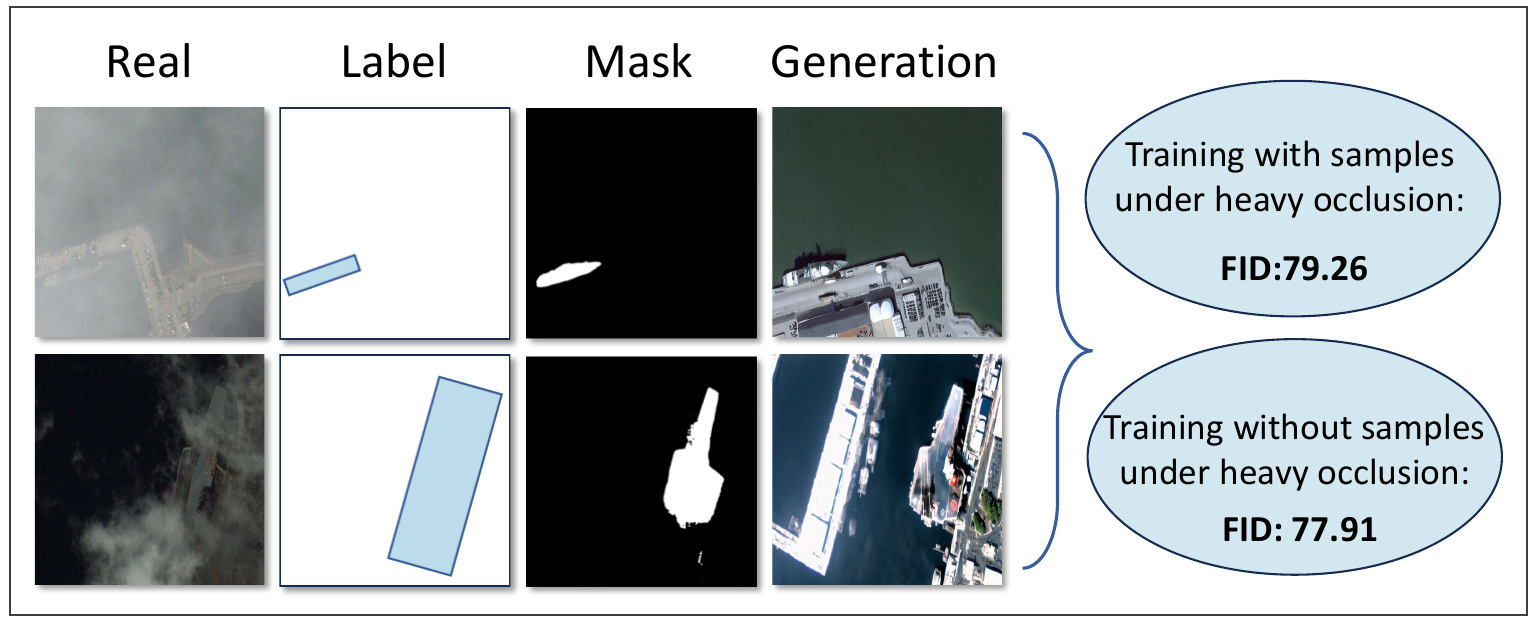}
  \caption{The results of OF-Diff in handling objects under heavy occlusion. The results indicate that while severe occlusion does indeed cause a certain degree of degradation in the quality of the target mask extracted by ESGM, it has little impact on the generated quality FID.}
  \label{fig:fail_mask}
\end{figure}

%% ============================================================================

\subsection{Analysis and Discussion}\label{discuss}

According to the current experimental results, adding the DDPO strategy does not simultaneously outperform previous results on all metrics. Using reinforcement learning strategies can indeed improve the performance of downstream tasks, but it does not necessarily improve the quality of image generation simultaneously. In other words, reinforcement learning strategies can also purposefully improve the quality of image generation, but this may come at the cost of not improving the performance of downstream tasks.

The proposed OF-Diff injects object shape masks extracted from the image layout as controllable conditions into the diffusion model, which effectively enhances object fidelity and improves the generation of small objects. However, this also makes the model dependent on the quality of the extracted shape masks. We analyze the impact of a distorted mask on the model's generated results. Specifically, we selected cases such as objects under heavy occlusion to examine the model's generation performance. Based on the analysis results in Figure\ref{fig:fail_mask}, we found that even under severe occlusion conditions, ESGM still demonstrates strong object mask extraction and generation capabilities. However, when the generated mask shape exhibits certain anomalies, it does produce objects matching that distorted shape. Nevertheless, this does not affect the overall FID and trainability of the generated images. Although the shapes we currently extract may exhibit edge anomalies in the object mask due to occlusion and other issues, complete errors are extremely rare.

\subsection{Quantitative Results on HRSC2016 Dataset}\label{hrsc2016_exp}
Table~\ref{tab:fid_map_hrsc2016} reports the comparative results on HRSC2016, where our method consistently achieves strong performance. Although it ranks second on CMMD, CAS, and YOLOScore—which mainly reflect aesthetic quality or local recognizability—it attains the best results on FID and KID, which measure distribution fidelity, as well as on the most crucial downstream metric, mAP50, outperforming the second-best method by +1.5\%. This indicates that our generated data preserves the real remote-sensing distribution more faithfully and thus provides more effective support for downstream tasks. A more detailed analysis is provided in Appendix~\ref{sup_aesthe}.
% ============================================
\begin{table}[h]
	\centering
	% 9pt is allowed
	\caption{Fidelity and Downstream Performance on HRSC2016}
	\label{tab:fid_map_hrsc2016}
        \fontsize{7}{10}\selectfont 

	% 'l''c', left center
        \setlength{\tabcolsep}{7pt} % 压缩列间距（默认6pt）
	\renewcommand{\arraystretch}{1.1}
	\begin{tabular}{l | cccc | cc}
		\Xhline{1pt} % head
		\multirow{2}{*}[-1pt]{\textbf{Method}} & \multicolumn{6}{c}{\textbf{HRSC216 Dataset}}\\ % [-6pt]手动调整上(+)下(-)
		\Xcline{2-7}{0.4pt} % 横线
        & FID$\downarrow$ & KID$\downarrow$ & CMMD$\downarrow$ & CAS$\uparrow$ & YOLOScore$\uparrow$ & mAP$_{50}\uparrow$ \\
		\Xcline{1-7}{0.4pt}
		LayoutDiff & 120.68 & 0.152 & 1.763 & 24.51 & 2.51 & 56.97 \\
		GLIGEN & 92.92 & 0.037 & 0.634 & 35.41 & 5.03 & 39.72 \\
		AeroGen & 97.44 & 0.055 & \textbf{0.51} & 39.62 & 16.4 & 47.68 \\
		CC-Diff & \underline{84.55} & \underline{0.035} & 0.681 & \textbf{45.27} & \textbf{32.42} & \underline{62.57} \\
		\Xcline{1-7}{0.4pt}
		\textbf{Ours} & \textbf{77.91} & \textbf{0.026} & \underline{0.573} & \underline{42.19} & \underline{30.97} & \textbf{64.1} \\
		\Xcline{1-7}{1pt}
	\end{tabular}
\end{table}

%% ============================================================================
\subsection{The mAP evolution given different amounts of synthetic and real data}\label{data_numbers}

\begin{table*}[t]
    \centering
    \caption{The impact of real and generated images at different ratios on mAP for downstream tasks.}
    \label{tab:data_numbers}
    \begin{tabular}{lc}
        \toprule
        \textbf{Data Composition} & \textbf{mAP (\%)} \\
        \midrule
        100\% Generated & 45.67 (-7.17) \\
        50\% Real + 50\% Generated & 50.74 (-2.10) \\
        100\% Real & 52.84 \\
        100\% Real + 50\% Generated & 53.92 (+1.08) \\
        100\% Real + 100\% Generated & 54.38 (+1.54) \\
        100\% Real + 200\% Generated & 54.74 (+1.90) \\
        100\% Real + 300\% Generated & 54.82 (+1.98) \\
        \bottomrule
    \end{tabular}
\end{table*}

We conduct multiple experiments on trainability using different quantities of real and generated data. The results are shown in the Table~\ref{tab:data_numbers}. Experimental results indicate that using only 100\% synthetic data struggles to achieve downstream task performance comparable to real data. However, this also demonstrates that even without a single real image, relying solely on synthetic images can enable object detection algorithms to achieve a mAP of 45.67\%. Furthermore, training with a larger volume of generated images can effectively enhance the model's object detection capabilities. However, when the amount of generated data reaches three times that of real data (based on the generation setting described in the paper), downstream performance shows little further improvement.

%% ============================================================================
\subsection{The Computational Cost}\label{cost}

\begin{table}[t]
    \centering
    \caption{The data on the computational cost of training OF-Diff compared to the key baselines.}
    \label{tab:cost}
    \begin{tabular}{lccc}
        \toprule
        \textbf{Models} & \textbf{\begin{tabular}[c]{@{}c@{}}Train GPU Mean\\Memory (MB)\end{tabular}} & \textbf{\begin{tabular}[c]{@{}c@{}}Train GPU\\Hours\end{tabular}} & \textbf{\begin{tabular}[c]{@{}c@{}}Inference Mean\\Time/Sample (s)\end{tabular}} \\
        \midrule
        LayoutDiff & 29232 & 41.33 & $<$1s \\
        GLIGEN & 14186 & 57.76 & 5.18 \\
        AeroGen & 27634 & 49.52 & 1.85 \\
        CC-Diff & 13668 & 38.01 & 3.96 \\
        OF-Diff & 27340 & 44.27 & 3.42 \\
        \bottomrule
    \end{tabular}
\end{table}

We provide the data on the computational cost of training OF-Diff compared to the key baselines in the Table~.\ref{tab:cost}. Experimental results indicate that although OF-Diff is not the most optimal in terms of training costs (GPU memory and GPU hours) and inference time, it remains nearly the second-best among these methods and does not incur high computational costs.

%% ============================================================================
\subsection{Aesthetic–downstream performance conflict}\label{sup_aesthe}
To further reveal the potential conflict between aesthetic quality and downstream task performance, we conduct a three-part analysis consisting of questionnaire evaluation, downstream performance comparison, and feature-level visualization.

\begin{figure}[t]
  \centering
  \includegraphics[width=0.95\columnwidth]{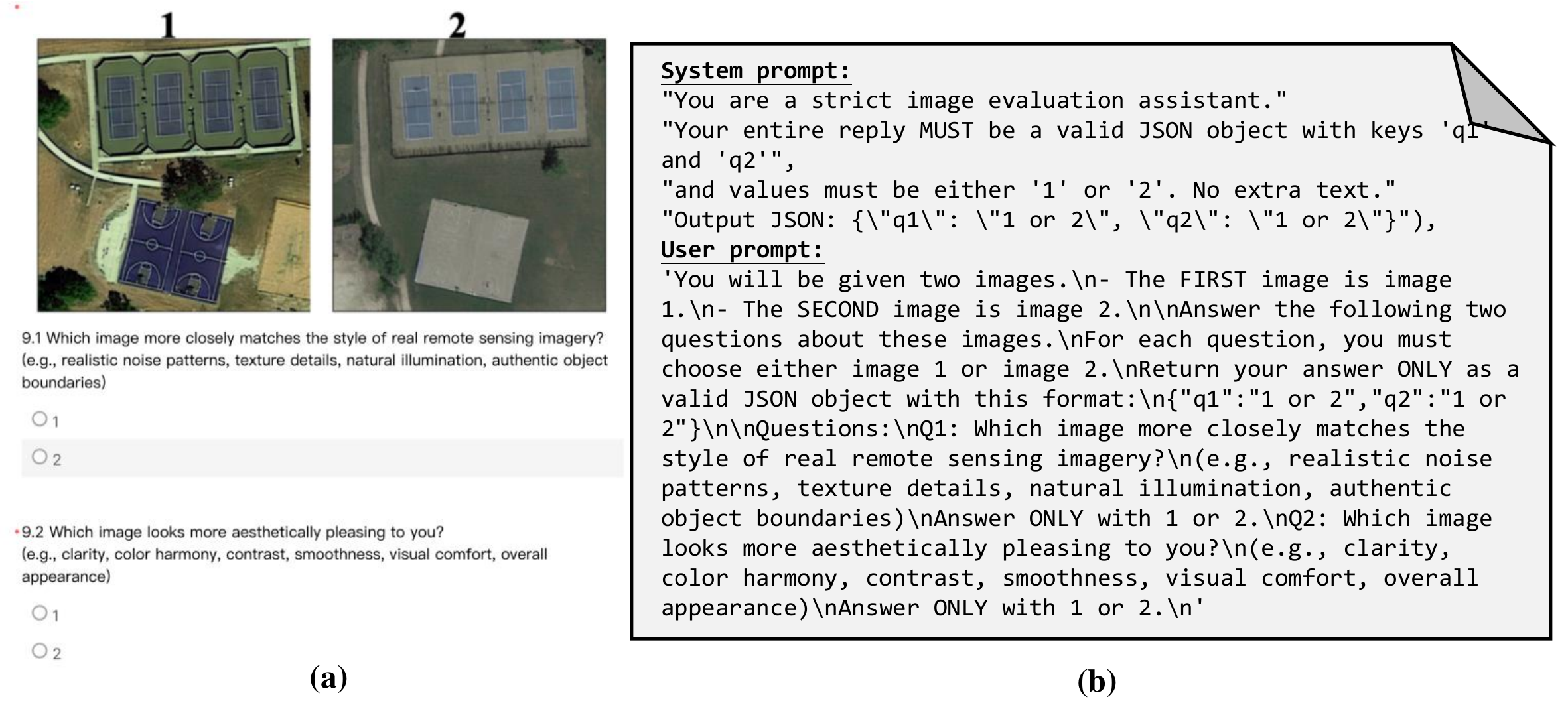}
  \caption{Aesthetic evaluation questionnaire design for generated images: (a) human experts, (b) GPT-5.}
  \label{fig:caption_questions}
\end{figure}

(1) \textbf{Human/GPT questionnaire study.}
As shown in Fig.~\ref{fig:caption_questions}, we design two targeted questions:

Q1. Which image more closely matches the style of real remote-sensing imagery?
(e.g., realistic noise patterns, texture details, natural illumination, authentic object boundaries)

Q2. Which image looks more aesthetically pleasing?
(e.g., clarity, color harmony, contrast, smoothness, visual comfort, overall appearance)

We invite 8 PhD researchers and 8 remote-sensing experts to participate, and additionally perform 3 rounds of GPT-5 evaluation. For each class in DIOR, we randomly sample one pair of images generated with vs. without captions (from the same ground truth), resulting in 20 image pairs. Each pair is randomly shuffled to avoid positional bias. The results are shown in Table~\ref{tab:aethe-results}, each value represents the average frequency with which the corresponding option was selected across all questionnaires.
\begin{table}[h]
	\centering
	% 9pt is allowed
	\caption{Single-choice results from human experts and GPT-5 (averaged over multiple annotators or repeated evaluations).}
	\label{tab:aethe-results}
        \fontsize{9}{11}\selectfont 
	% 'l''c', left center
        \setlength{\tabcolsep}{8pt} % 压缩列间距（默认6pt）
	\renewcommand{\arraystretch}{1.2}
	\begin{tabular}{l | cc | cc}
		\Xhline{1pt} % head
		\multirow{2}{*}[-1pt]{\textbf{Option}} & \multicolumn{2}{c|}{\textbf{Human experts}} & \multicolumn{2}{c}{\textbf{GPT-5}} \\ % [-6pt]手动调整上(+)下(-)
		\Xcline{2-5}{0.4pt} % 横线
		& Q1 & Q2 & Q1 & Q2 \\
		\Xcline{1-5}{0.4pt}
		w./ caption & 6.57 & \textbf{11.21} & 2.33 & \textbf{15.33} \\
		w./o. caption & \textbf{13.43} & 8.79 & \textbf{17.67} & 4.67 \\
		\Xcline{1-5}{1pt}
	\end{tabular}
\end{table}

Both human experts and GPT consistently prefer the caption-conditioned images in terms of aesthetics, but find them less similar to real remote-sensing imagery. In contrast, images generated without captions appear less visually appealing but better preserve the real-world texture and structural characteristics needed for downstream tasks.

\begin{figure}[t]
  \centering
  \includegraphics[width=0.8\columnwidth]{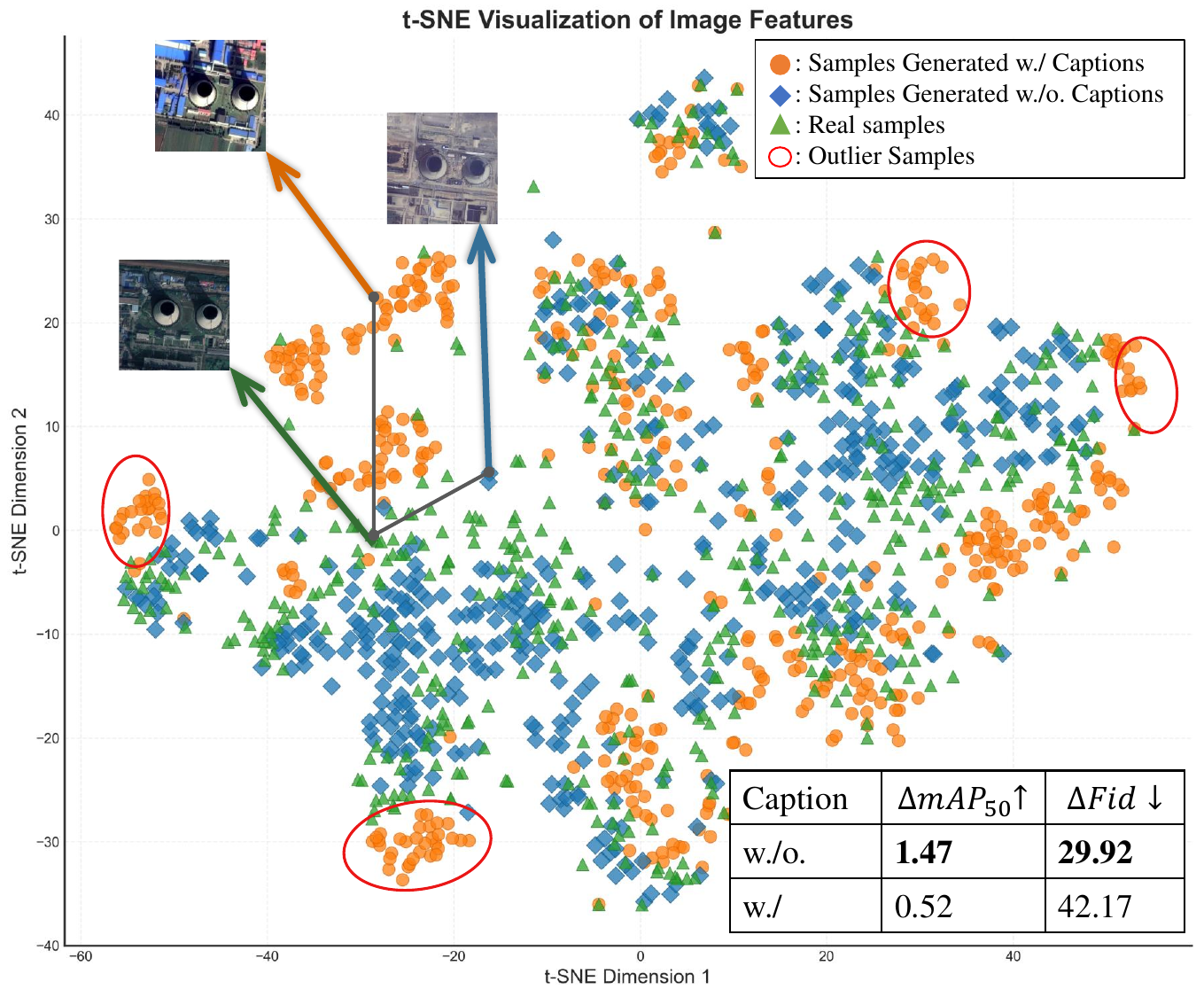}
  \caption{t-SNE feature analysis of generated samples w./ and w./o. captions. Incorporating captions produces a large number of outlier samples, lowers fidelity (higher FID), and further degrades downstream performance (lower $\Delta{mAP_{50}}$).}
  \label{fig:tsne-caption}
\end{figure}

\textbf{(2) Downstream performance comparison.}
On the DIOR dataset, as shown in the table at the bottom-right of Fig.~\ref{fig:tsne-caption}, adding captions reduces the downstream improvement $\Delta\text{mAP}_{50}$ by 1.15 and also leads to a significantly higher FID. Combined with finding (1), this reveals that caption-guided generation tends to over-beautify images—masking the natural imperfections of remote-sensing imagery—and consequently harms downstream performance.

\textbf{(3) Feature-level visualization.}
We also visualize features using t-SNE in Fig.~\ref{fig:tsne-caption}. We observe that adding captions produces more outliers, whereas samples generated without captions align more closely with the GT distribution, indicating higher fidelity.

\textbf{Taken together}, these findings suggest that models should remain faithful to the inherent quirks and imperfections of the original remote-sensing data, rather than generating overly “idealized’’ or aesthetically enhanced imagery. Incorporating captions risks amplifying the latter behavior.

\begin{figure}[t]
  \centering
  \includegraphics[width=0.8\columnwidth]{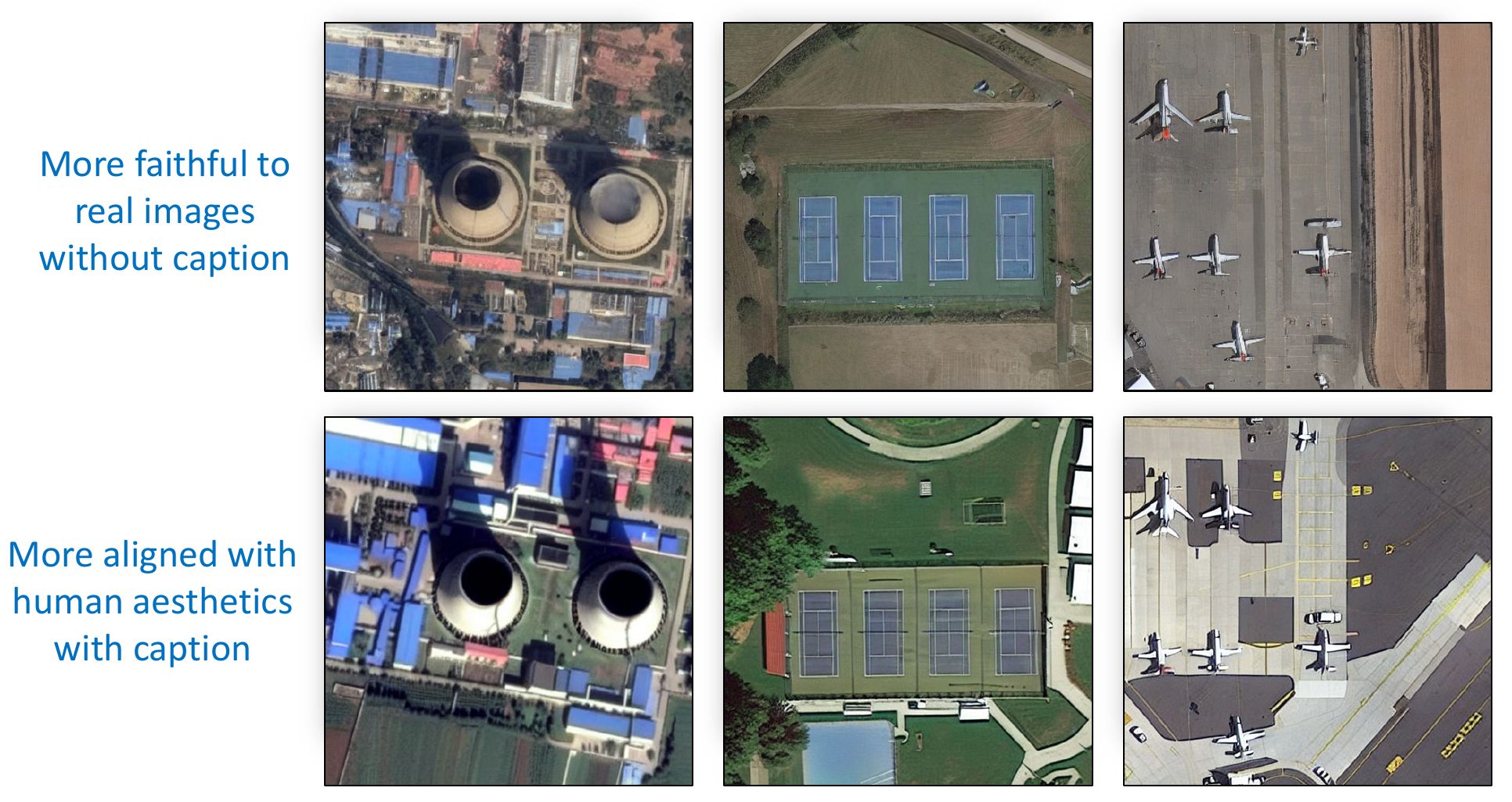}
  \caption{The influence of caption on the generation of images in terms of being more realistic and more aesthetically pleasing.}
  \label{fig:w_o_caption}
\end{figure}

Additional examples are provided in Fig.~\ref{fig:w_o_caption}.
%% ============================================================================

%% =============================================================================

\subsection{More Qualitative and Quantitative Results}\label{more_results}

%% =============================================================================
%% Trainability
\begin{table}[h]
    \centering
   
    \caption{Trainability ($\uparrow$) comparison on DIOR and DOTA. `Baseline' denotes accuracy with the unaugmented dataset.}
    \label{tab:mAP_results}
    
    %–– 缩小整体字号（比 9pt 更紧凑）
    \small
    %–– 缩短列与列之间的空白（默认 6pt）
    \setlength{\tabcolsep}{10pt}
    %–– 行高
    \renewcommand{\arraystretch}{1.2}

    %\resizebox{\linewidth}{!}{%
    \begin{tabular}{l|ccc|ccc}
        \Xhline{1pt}
        % Method 的 \multirow 不再额外位移，并显式水平居中
        \multirow{2}{*}{\centering\textbf{Method}} 
        & \multicolumn{3}{c|}{\textbf{DIOR Dataset}} 
        & \multicolumn{3}{c}{\textbf{DOTA Dataset}} \\ 
        \Xcline{2-7}{0.4pt}
        & mAP & mAP$_{50}$ & mAP$_{75}$
        & mAP & mAP$_{50}$ & mAP$_{75}$ \\
        \Xcline{1-7}{0.4pt}
        Baseline & 30.51 & 52.84  & 32.10 & 38.09 & 66.31 & 38.44 \\
        LayoutDiff & 29.81 & 52.14  & 30.36 & 38.91 & 66.75 & 40.37 \\
        GLIGEN     & 28.48 & 51.27  & 29.21 & 38.84 & 66.10 & 40.24 \\
        AeroGen    & 31.53 & 53.37 & 33.60 & 38.45 & 67.09 & 39.07 \\
        CC-Diff    & 31.82 & 53.48 & 33.97 & 38.51 & 66.52 & 39.02 \\
        \Xcline{1-7}{0.4pt}
        \textbf{Ours} & \textbf{32.71} & \textbf{54.44} & \textbf{34.05} & \textbf{40.03} & \textbf{67.89} & \textbf{42.20} \\
        \Xcline{1-7}{1pt}
    \end{tabular}
    %} % 取消上一行 \resizebox 时，同时把这一行的百分号删除
\end{table}

% ============================================
\begin{figure}[t]
  \centering % 

  % (DIOR)
  \includegraphics[width=0.95\textwidth]{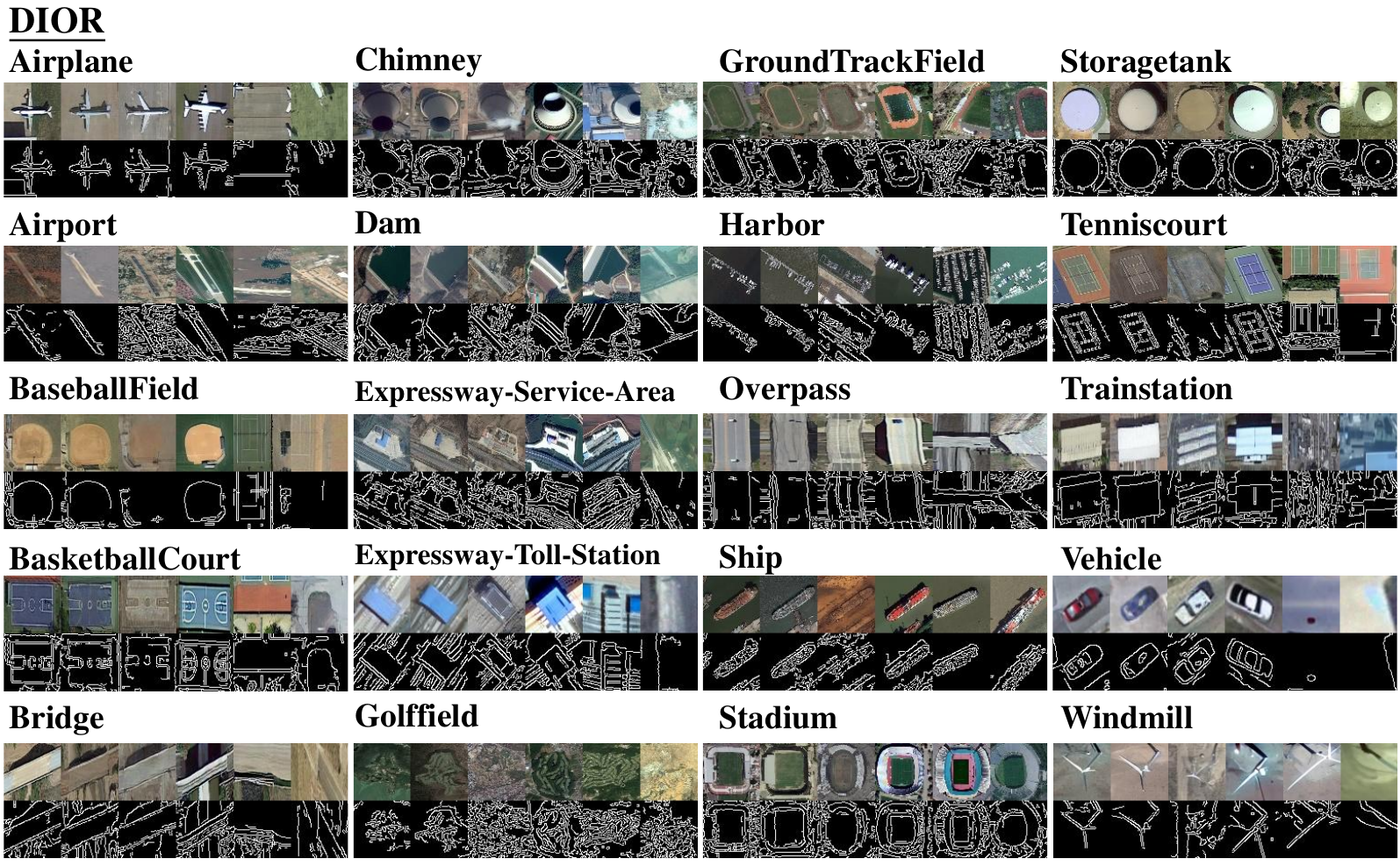}
  %\captionlistentry{DIOR Results} 
  \label{fig:sup_edge_vis_dior_sub}

  \vspace{3mm} % 

  % (DOTA)
  \includegraphics[width=0.95\textwidth]{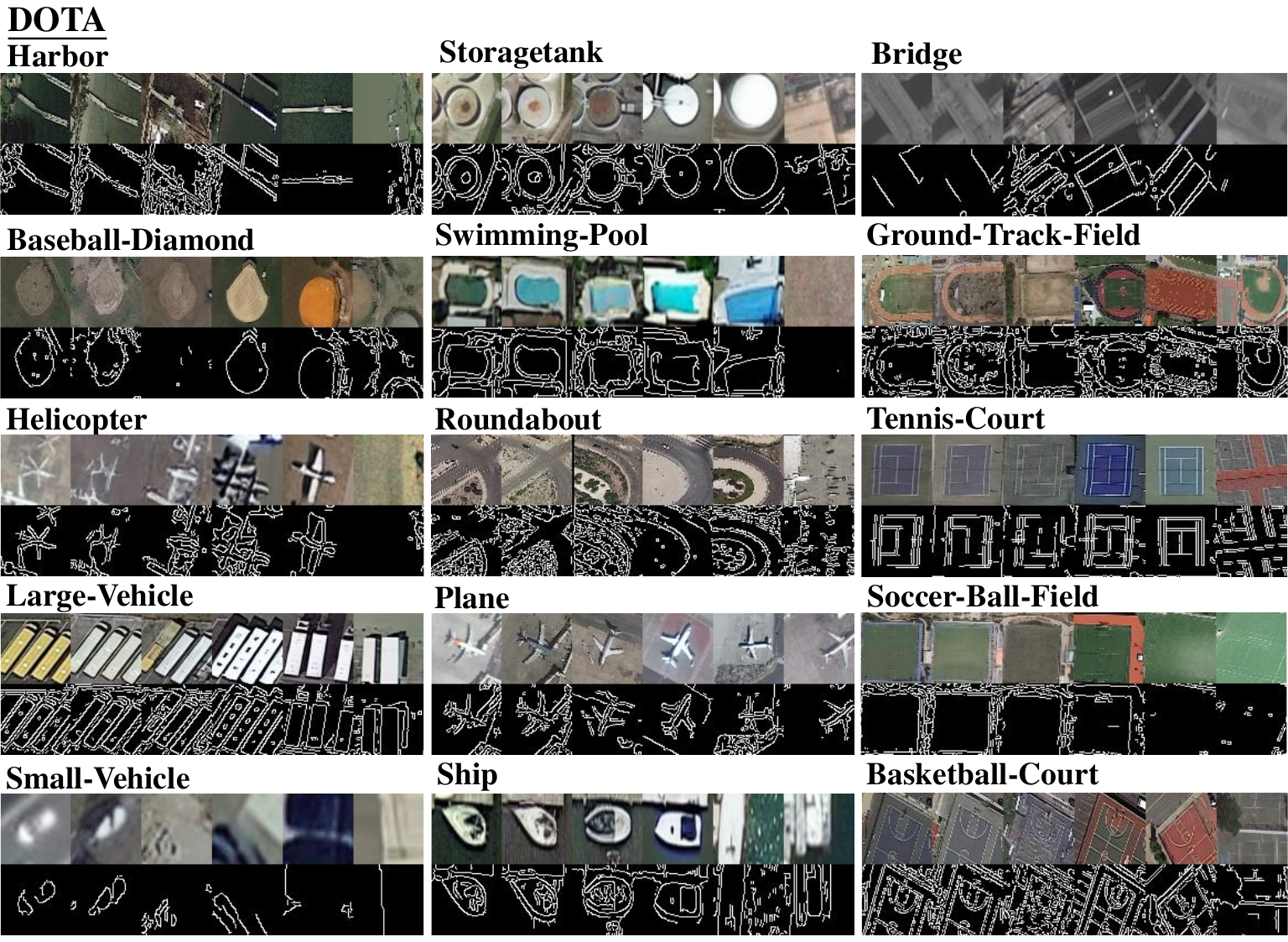}
  %\captionlistentry{DOTA Results}
  \label{fig:sup_edge_vis_dota_sub}

  \caption{Comparison of generated instance patches and their Canny edge maps for the same bbox on the DIOR and DOTA dataset. Each image set is ordered as follows: Ground Truth, OF-Diff, AeroGen, CC-Diff, GLIGEN, and LayoutDiff.}
  \label{fig:sup_edge_vis_combined}
\end{figure}

\clearpage

% APs for DIOR
\begin{table}[t]
    \centering
    \fontsize{8}{10}\selectfont
    \setlength{\tabcolsep}{3.2pt} % 压缩列间距（默认6pt）
    \renewcommand{\arraystretch}{1.5}

    %=======================  (Part 1)  =========================%
    \begin{tabular}{l | ccccc ccccc}
        \Xhline{1pt}
        \multirow{2}{*}[-6.5pt]{\textbf{Method}} & \multicolumn{9}{c}{\textbf{DIOR Dataset}} \\ 
        \Xcline{2-11}{0.4pt}
        & \makecell{Expressway \\ Service-area}  & \makecell{Expressway \\ toll-station} & Airplane & Airport & \makecell{Baseball \\ field} & \makecell{Basketball \\ court} & Bridge & Chimney & Dam & \makecell{Golf \\ Field} \\
        \Xcline{1-11}{0.4pt}
        LayoutDiff  & 53.1 & 44.8 & 62.8 & 29.4 & 63.2 & 79.6 & 25.9 & 72.6 & 22.4 & 69.3 \\
        GLIGEN      & 52.7 & 44.8 & 62.6 & 26.7 & 63.0 & 79.6 & 25.2 & 72.6 & 19.5 & 67.4 \\
        AeroGen     & \textbf{58.1} & \textbf{45.2} & 63.1 & 32.7 & \textbf{63.4} & \textbf{81.0} & 29.5 & 72.6 & 21.1 & 69.1 \\
        CC-Diff     & 53.5 & 45.1 & 62.9 & \textbf{38.4} & 63.3 & 79.9 & 29.3 & \textbf{72.7} & \textbf{27.6} & 70.5 \\
        \textbf{Ours}& 58.0 & 44.9 & \textbf{71.3} & 37.0 & 63.2 & 80.2 & \textbf{30.1} & 72.5 & 24.9  & \textbf{75.4}\\
        \Xcline{1-11}{1pt}
    \end{tabular}

    \vspace{8pt} % ← 两个 tabular 之间留空，可按需调整

    %=======================  (Part 2)  =========================%
    \begin{tabular}{l | ccccc ccccc}
        \Xhline{1pt}
        \multirow{2}{*}[-6.5pt]{\textbf{Method}} & \multicolumn{10}{c}{\textbf{DIOR Dataset}} \\ 
        \Xcline{2-11}{0.4pt}
        & \makecell{Ground \\ Track-field} & Harbor & Overpass & Ship & Stadium & \makecell{Storage \\ Tank} & \makecell{Tennis \\ Court} & Trainstation & Vehicle & Windmill \\
        \Xcline{1-11}{0.4pt}
        LayoutDiff  & \textbf{71.2} & 32.8 & 43.9 & 62.9 & 59.0 & \textbf{52.5} & 72.4 & 52.1 & 26.9 & 46.0 \\
        GLIGEN      & 70.1 & 30.3 & 45.8 & 62.8 & 56.8 & 52.0 & \textbf{72.5} & 49.0 & 26.9 & 45.3 \\
        AeroGen     & 71.0 & 42.7 & \textbf{50.7} & 62.9 & 56.6 & 44.5 & \textbf{72.5} & 52.6 & \textbf{31.4} & \textbf{46.7} \\
        CC-Diff     & 64.6 & 43.1 & 49.0 & 63.0 & \textbf{61.7} & 44.7 & 72.4 & \textbf{54.4} & 27.0 & 46.5 \\
        \textbf{Ours}& 66.3 & \textbf{43.9} & 49.4 & \textbf{70.5} & 52.7 & 44.4 & 72.4 & 54.1 & 31.0 & \textbf{46.7} \\
        \Xcline{1-11}{1pt}
    \end{tabular}

    %===================  统一 caption 与 label  =================%
    \caption{Detailed downstream trainability results (measured by average precision) on the DIOR dataset.}
    \label{tab:DIOR-APs}
\end{table}

% APs for DOTA
\begin{table}[t]
    \centering
    \fontsize{8}{10}\selectfont
    \setlength{\tabcolsep}{3.2pt} % 压缩列间距（默认6pt）
    \renewcommand{\arraystretch}{1.5}

    %============  (Part 1) – 9 columns  ============%
    \begin{tabular}{l | cccccccc}      % 1(方法) + 8(类别) = 9 列
      \Xhline{1pt}
      \multirow{2}{*}[-6pt]{\textbf{Method}}
         & \multicolumn{7}{c}{\textbf{DOTA Dataset}} \\   % ← 8 列
      \Xcline{2-9}{0.4pt}
         & \makecell{Plane}
         & \makecell{Baseball-diamond}
         & Bridge & \makecell{Ground \\ Track-field} & Small-vehicle
         & \makecell{Large-vehicle}
         & \makecell{Ship}
         & Tennis-court \\
      \Xcline{1-9}{0.4pt}
      LayoutDiff   & 80.4 & 74.2 & \textbf{48.8} & 59.9 & 62.9 & 72.7 & 82.5 & 89.6  \\
      GLIGEN       & 87.0 & 72.9 & 47.3 & 56.4 & 63.7 & 73.1 & 82.7 & 90.1  \\
      AeroGen      & 86.1 & \textbf{77.3} & 48.6 & 58.6 & 64.5 & \textbf{78.1} & 82.5 & 83.3  \\
      CC-Diff      & \textbf{87.2} & 73.4 & 47.1 & 57.8 & 64.3 & 73.9 & 82.6 & 89.2  \\
      \textbf{Ours}& 85.2 & 75.3 & 46.5 & \textbf{60.4} & \textbf{68.3} & 77.2 & \textbf{84.4} & \textbf{90.4} \\
      \Xcline{1-9}{1pt}
    \end{tabular}

    \vspace{8pt} % ← 两个 tabular 之间留空，可按需调整

    %============  (Part 2) – 8 columns  ============%
    \begin{tabular}{l | ccccccc}      % 1(方法) + 7(类别) = 8 列
      \Xhline{1pt}
      \multirow{2}{*}[-6pt]{\textbf{Method}}
         & \multicolumn{7}{c}{\textbf{DOTA Dataset}} \\   % ← 7 列
      \Xcline{2-8}{0.4pt}
         & \makecell{Basketball-court}
         & \makecell{Storage-tank}
         & Soccer-ball-field & Roundabout
         & \makecell{Harbor}
         & \makecell{Swimming-pool}
         & Helicopter \\
      \Xcline{1-8}{0.4pt}
      LayoutDiff   & 78.9 & 76.3 & \textbf{46.3} & \textbf{47.6} & 60.9 & 62.1 & \textbf{57.9} \\
      GLIGEN       & 79.6 & 76.5 & 42.2 & 43.1 & 60.7 & 62.3 & 53.8 \\
      AeroGen      & 77.3 & 79.9 & 44.8 & 46.6 & 59.4 & 62.6 & 56.4 \\
      CC-Diff      & 79.0 & \textbf{82.7} & 42.7 & 43.1 & 58.9 & 62.7 & 52.9 \\
      \textbf{Ours}& \textbf{83.3} & 77.1 & 42.1 & 44.7 & \textbf{62.1} & \textbf{67.9} & 53.3\\
      \Xcline{1-8}{1pt}
    \end{tabular}

    %===================  统一 caption 与 label  =================%
    \caption{Detailed downstream trainability results (measured by average precision) on the DOTA dataset.}
    \label{tab:DOTA-APs}
\end{table}

% \clearpage

\begin{figure}[t]
  \centering
  \includegraphics[width=\textwidth]{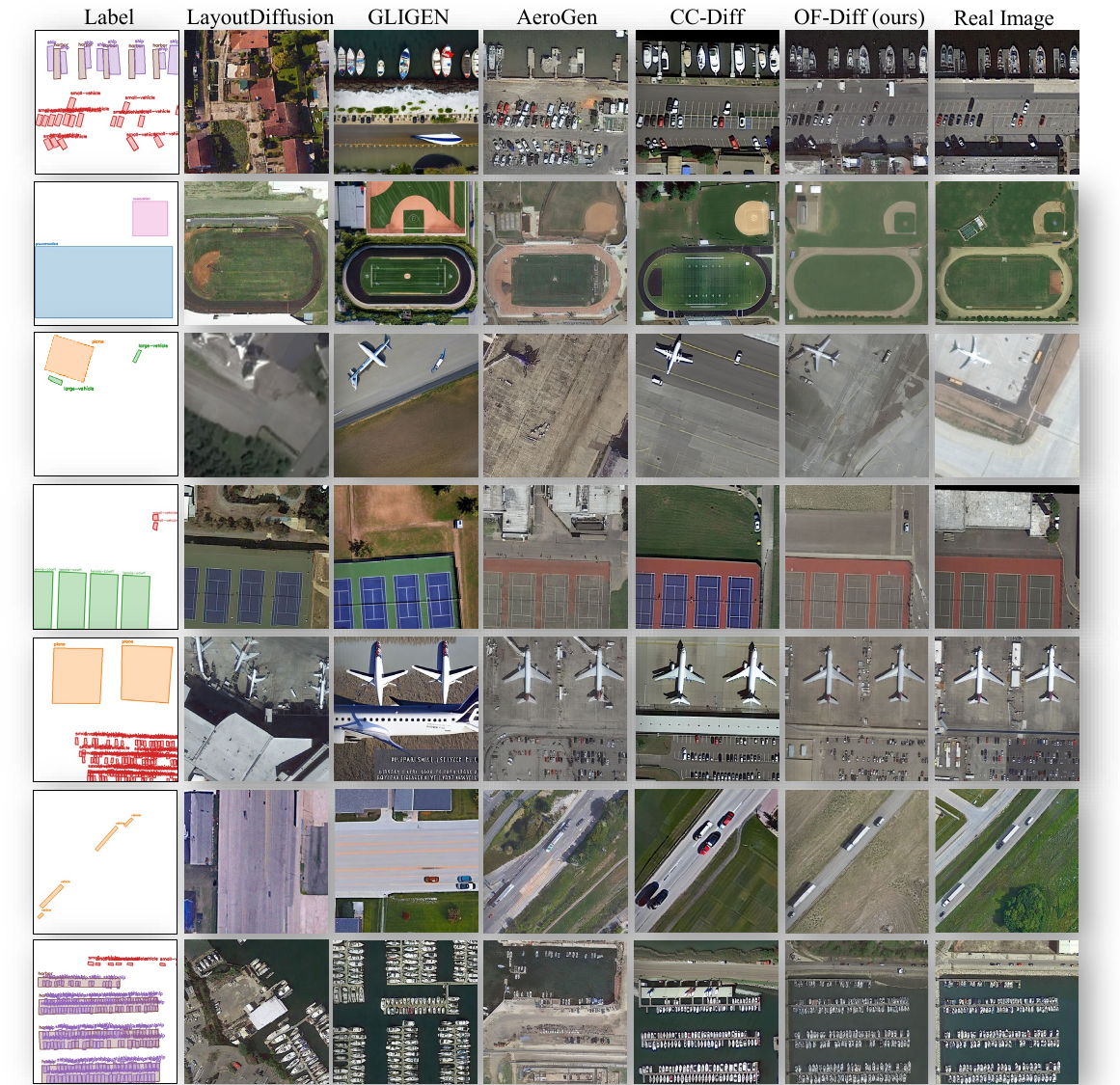}
  \caption{Additional qualitative results on DIOR and DOTA. The results demonstrate that OF-Diff has certain superiority and accuracy in generating small objects, and it also has an advantage in generating the shapes of objects. For instance, the aircraft target in the third row is generated more accurately by OF-Diff, with a more realistic structure. The small vehicles in the fourth and fifth rows and the large vehicle in the sixth row are also more accurately generated. Additionally, the small ship in the seventh row is generated with greater accuracy.}
  \label{fig:ext_dota}
\end{figure}

\end{document}